\documentclass[10pt,twocolumn,letterpaper]{article}

\PassOptionsToPackage{table}{xcolor}
\usepackage{iccv}              %

\usepackage{array}
\usepackage{algorithm}
\usepackage{algpseudocode}
\usepackage{amsmath}
\usepackage{booktabs}
\usepackage{tabularx}
\usepackage{adjustbox}
\usepackage{multirow}
\usepackage{graphicx}
\usepackage{float}
\usepackage[table]{xcolor}
\usepackage{colortbl}
\usepackage{makecell}
\usepackage{caption}
\usepackage{relsize}
\usepackage{marvosym}

\definecolor{lightyellow}{RGB}{255,255,200}
\definecolor{lightorange}{RGB}{255,230,200}
\definecolor{lightred}{RGB}{255,200,200}

\newcommand{\tbest}[1]{\cellcolor{lightyellow}#1}
\newcommand{\sbest}[1]{\cellcolor{lightorange}#1}
\newcommand{\best}[1]{\cellcolor{lightred}#1}

\newcommand{\cL}{\mathcal{L}}

\DeclareMathOperator*{\argmin}{argmin~}

\makeatletter
\DeclareRobustCommand\onedot{\futurelet\@let@token\@onedot}
\def\@onedot{\ifx\@let@token.\else.\null\fi\xspace}

\makeatother

\renewcommand{\eqref}[1]{Eq.~\ref{#1}}

\newcommand{\boldparagraph}[1]{\vspace{0.2cm}\noindent{\bf #1:} }

\newif\ifcomment
\commenttrue
\ifcomment
	\newcommand{\yl}[1]{ \noindent {\color{cyan} {\bf YL:} {#1}} }
	
\else
	\newcommand{\ag}[1]{}
	\newcommand{\yl}[1]{}
\fi

\newcommand{\method}[1]{Vivid4D}{}

\definecolor{MyGreen}{rgb}{0.02,0.4,0.02}
\definecolor{MyRed}{rgb}{0.9,0.1,0.1}

\definecolor{iccvblue}{rgb}{0.21,0.49,0.74}

\usepackage[pagebackref,breaklinks,colorlinks,allcolors=iccvblue]{hyperref}

\def\confName{ICCV}
\def\confYear{2025}

\title{Vivid4D: Improving 4D Reconstruction from Monocular Video by \\ Video Inpainting}
\author{
Jiaxin Huang$^{1}$,
Sheng Miao$^{1}$,
BangBang Yang$^{2}$,
Yuewen Ma$^{2}$,
Yiyi Liao$^{1}\textsuperscript{\Letter}$
\vspace{0.2cm}
\\
{\normalsize $^{1}$ Zhejiang University\quad$^{2}$ ByteDance}
\\
\small{Project Page: \url{https://xdimlab.github.io/Vivid4D/}}
}

\begin{document}

\setlength{\abovedisplayskip}{0.2 em}
\setlength{\belowdisplayskip}{0.2 em}

\twocolumn[{%
	\renewcommand\twocolumn[1][]{#1}%
	\maketitle
		\vspace{-2.5em}
	\begin{center}
		\includegraphics[width=0.95\textwidth]{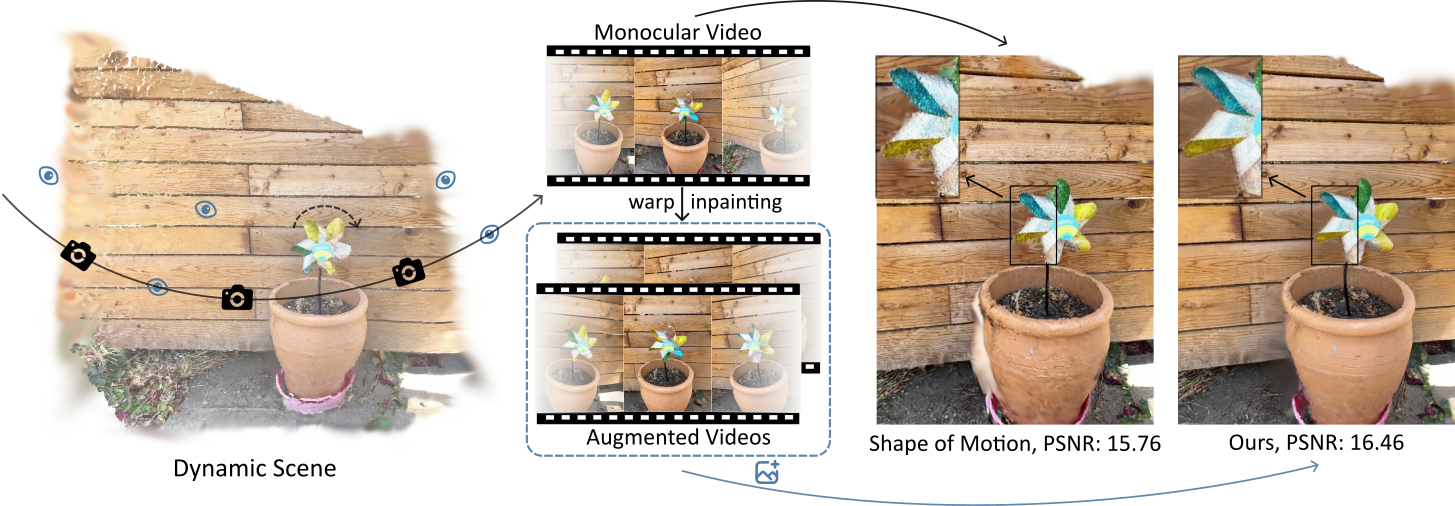}
\vspace{-0.2cm}			
   \captionsetup{type=figure}
			\captionof{figure}{
				\textbf{
					\method{}}. We improve dynamic scene reconstruction from casually captured monocular videos by synthesizing augmented views. Our approach integrates both geometric and generative priors to reformulate the video augmentation as a video inpainting task.
                    This enables our method to effectively complete invisible regions in the scene and enhance reconstruction quality.
			}
			\label{fig: teaser}
		\end{center}
}]
\vspace{0.35cm}

\begin{abstract}
Reconstructing 4D dynamic scenes from casually captured monocular videos is valuable but highly challenging, as each timestamp is observed from a single viewpoint. We introduce \method{}, a novel approach that enhances 4D monocular video synthesis by augmenting observation views — synthesizing multi-view videos from a monocular input. Unlike existing methods that either solely leverage geometric priors for supervision or use generative priors while overlooking geometry, we integrate both. This reformulates view augmentation as a video inpainting task, where observed views are warped into new viewpoints based on monocular depth priors. To achieve this, we train a video inpainting model on unposed web videos with synthetically generated masks that mimic warping occlusions, ensuring spatially and temporally consistent completion of missing regions. To further mitigate inaccuracies in monocular depth priors, we introduce an iterative view augmentation strategy and a robust reconstruction loss. Experiments demonstrate that our method effectively improves monocular 4D scene reconstruction and completion.

\end{abstract}
    
\section{Introduction}
\label{sec: intro}

Reconstructing dynamic 3D scenes from convenient monocular video captures is a highly valuable task in computer vision and graphics, yet remains a profound challenge.
Recent advances in 4D reconstruction~\cite{4dgs_deformation, 4dgs_rotor, 4dgs_time, dnerf, hexplane, fastdnerf, kplane, dynamicgs, gaussianflow} have made significant progress in modeling dynamic scenes, but they typically rely on synchronized multi-view cameras. 

Given the sparse observations of monocular videos where each timestamp is observed from a single viewpoint, existing methods attempt to address this challenging task by incorporating additional priors.
One line of studies has explored various geometric priors for monocular 4D reconstruction, including geometric constraints~\cite{nerfies, hypernerf}, optical flow~\cite{neuralsceneflow, dynibar, gflow}, depth estimation~\cite{modgs, dynamicviewsynthesis}, and tracking~\cite{mosca, som, marble}. However, existing methods yield suboptimal results for two main reasons.
First, these auxiliary supervisions are not consistently reliable and may not correlate linearly with rendering performance; for example, a small depth error may lead to significant color shifts in regions with texture changes. %
Second, these priors, which are derived solely from input views, provide limited guidance for regions occluded or unobserved in the input views. 
More recently, video diffusion models have emerged as another type of prior for such tasks, owing to their impressive ability to generate realistic videos~\cite{svd, sora, cogvideox, hunyuanvideo}. These models can generate plausible RGB images for unknown viewpoints, overcoming the disadvantages of using geometry priors alone. %
However, existing methods are either limited to static scenes~\cite{reconfusion, viewcrafter, see3d, sgd, viewextrapolator}, or require a large amount of posed images for training~\cite{cat4d, genxd}, while collecting such training data is highly challenging as dynamic scenes' camera poses are difficult to obtain.

\begin{figure}[t]
    \centering
    \includegraphics[width=\columnwidth]{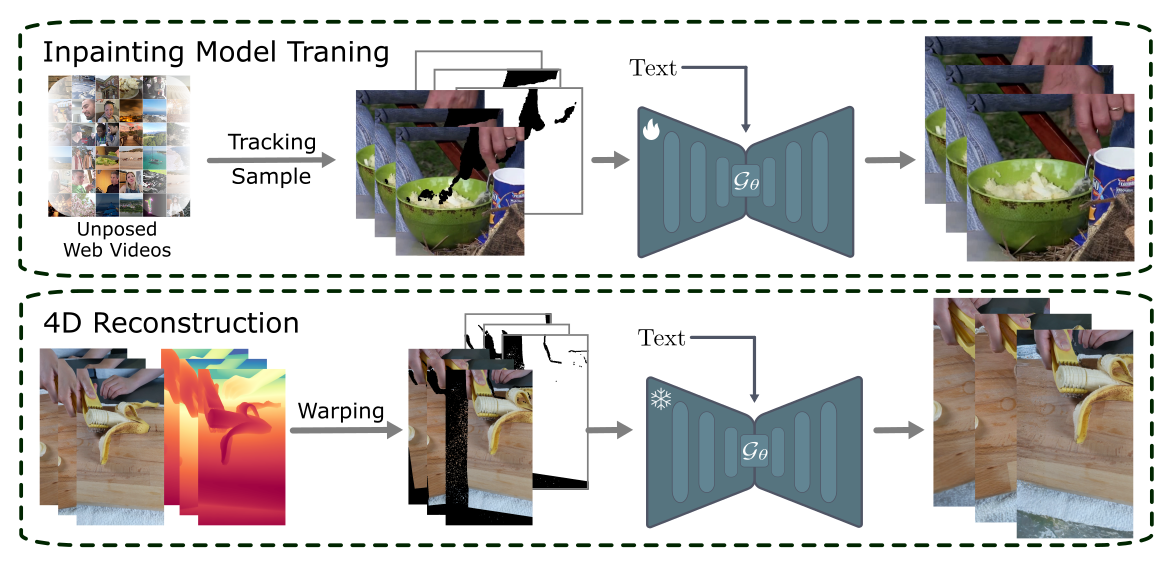}
    \vspace{-0.6cm}
    \caption{\textbf{Video Inpainting for 4D Reconstruction}. To train the video inpainting model, we use 2D tracking to generate masked training pairs from  \textit{unposed} web videos. During 4D reconstruction, we warp the monocular video to novel viewpoints, creating masked videos that our inpainting diffusion model then completes.}
   \label{fig: diffusion_illustration}
   \vspace{-0.3cm}
\end{figure}

In this work, we propose Vivid4D, a novel approach for 4D monocular reconstruction that integrates geometric priors from input videos with the generative capabilities of video diffusion models. Instead of directly generating novel views, we reformulate view augmentation as a video inpainting task: first, observed views are warped to unseen viewpoints based on geometric priors; then, a video diffusion model inpaints missing regions to ensure spatial-temporal consistency.
This allows us to train the video inpainting model using \textit{unposed} web videos by mimicking warping occlusions based on 2D trackings. 
Specifically, our method includes two key steps as shown in~\cref{fig: diffusion_illustration}.
First, we train a video inpainting model tailored for our view augmentation task that learns to in-paint occluded regions in a spatial-temporally coherent manner (\cref{fig: diffusion_illustration} top). This model is trained on unposed web videos, using 2D trackings to synthesize occlusion masks.
Next, we use the trained video inpainting model for view augmentation in monocular-video-based 4D reconstruction, leveraging monocular depth priors to warp the video to unseen viewpoints, which our inpainting model then completes (\cref{fig: diffusion_illustration} bottom).
To maximize scene coverage while compensating for inaccurate monocular depth priors, we propose a progressive view expansion strategy that iteratively generates augmented views, effectively accumulating comprehensive multi-view observations.
When further combined with a robust reconstruction loss, the augmented views
enables \method{} to improve 4D reconstruction while maintaining geometric consistency and temporal coherence.

The main contributions of our work can be summarized as follows:
    1) We propose \method{}, a novel method for monocular 4D scene reconstruction that leverages multi-view supervision from both video diffusion models and depth-guided warping.
    2) We reformulate view augmentation as a video inpainting task, enabling efficient training on unposed web videos through synthesized occlusion masks based on 2D tracking.
    3) We design an iterative view augmentation strategy to progressively expand the observation viewpoint of the scene, enabling robust reconstruction and novel view synthesis.
    4) We conduct extensive experiments across various dynamic scenes and show that our method improves the quality of novel view synthesis compared to existing approaches on different types of motion and scene complexity.

\section{Related Work}
\label{sec: related_work}

\paragraph{\textbf{3D Novel View Synthesis.}} 
Recent advances in neural rendering, from NeRF~\cite{nerf, mipnerf, mipnerf360, zipnerf, edus, f2nerf} to 3D-GS~\cite{3dgs, hugs, streetgs, 2dgs, mipsplatting, sugar, pixelsplat, scaffoldgs}, have revolutionized static scene reconstruction and novel view synthesis. 
To improve reconstruction quality, especially in challenging scenarios such as sparse inputs, researchers have explored two main directions: integrating traditional geometric reconstruction techniques~\cite{instantsplat, corgs, mvsplat} and applying geometric regularizations~\cite{freenerf, dnerf, regnerf, sparsenerf, dngaussian, da4nerf}. Recently, generative models have been leveraged for additional visual supervision. For instance, several methods~\cite{sparsefusion, sparsegs, svsgs} employ score distillation sampling (SDS)~\cite{dreamfusion} to distill knowledge from pre-trained diffusion models, optimizing the 3D representation to align with learned priors. Another line of work~\cite{reconfusion, see3d, viewcrafter, lmgs, sp2360, viewextrapolator, sgd, genwarp} projects dense scene representations~\cite{dust3r, nerf, 3dgs, genwarp} and projects them into novel viewpoints, guiding reconstruction with geometrically consistent views generated by diffusion models.
Meanwhile, \cite{reconx} incorporates 3D structure conditioning into video diffusion for 3D-consistent video generation, and~\cite{cat3d} generates multi-view images conditioned on input views and target camera poses to provide additional supervision. 
While these methods achieve impressive results, they are typically limited to static scenes and struggle with dynamic objects and temporal consistency.
 
\paragraph{\textbf{4D Monocular Novel View Synthesis.}} 
Many methods have extended neural rendering to dynamic scenes, typically based on canonical space modeling with time-dependent deformation fields~\cite{4dgs_deformation, dnerf, fastdnerf, modgs, gaussianflow, dynamicgs} or 4D time-varying representations~\cite{space-timenerf,4dgs_time, 4dgs_rotor, l4gm,kplane, hexplane}. However, these methods typically require multi-view synchronized and calibrated videos, making them ill-posed when multi-view data is limited.

For monocular video reconstruction, some methods leverage data-driven priors like depth~\cite{depthanything, chronodepth}, tracks~\cite{trackanything}, and geometric constraints~\cite{som, mosca, gflow, marble, hypernerf, dynibar, modgs} for initialization or supervision, but these priors may not be fully accurate and do not always correlate linearly with RGB performance.
More recently, diffusion models have been introduced for 4D reconstruction. 
For example,~\cite{dreamscene4d} utilizes pre-trained diffusion models~\cite{sdinpaint} for scene completion, though the visual quality still requires improvement. Concurrent work~\cite{ar4d} uses LGM~\cite{lgm} to generate pseudo novel views as additional supervision, but it is primarily applied at object level. Other methods~\cite{cat4d, dimensionx, genxd} generate spatial-temporally consistent frames to optimize 4D scenes, but they require temporally and spatially decoupled data to learn disentangled control over camera motion and time. Additionally,~\cite{cat4d, genxd} rely on camera motion as input for training, which can be challenging to obtain for dynamic video datasets.
In contrast, our approach only requires unposed web videos with simulated warping occlusion masks, enabling training on a wider range of casual video datasets without camera pose information.
Moreover,~\cite{dimensionx} lacks precise camera control during inference, while our method implicitly incorporates precise pose information through warping.

\paragraph{\textbf{Video Diffusion Models.}} 
Video diffusion models have shown remarkable capabilities in video generation tasks. Recent works focued on video in-painting~\cite{cococo, diffueraser, floed, avid, fffvdi} and out-painting~\cite{m3ddm, motia, followyourcanvas} have demonstrated the effectiveness of diffusion models in generating temporally consistent content for incomplete videos. However, these methods mainly deal with a single video without perspective changes. Other generative methods~\cite{syncammaster, gcd} take camera poses as input to generate synchronized videos from novel viewpoints, but their quality needs improvement, and they are constrained by static input-output camera viewpoints.

Some works~\cite{stereocrafter, gs-dit} attempt to overcome viewpoint constraints by inpainting videos after perspective changes. These methods utilize geometric constraints~\cite{stereocrafter, gen3c, trajectorycrafter} or pseudo 4D Gaussian fields~\cite{gs-dit} to obtain masked videos after warping, and fine-tune video diffusion models~\cite{svd, cogvideo, cogvideox} for multi-view synthesis. While these approaches show the potential of diffusion models in generating consistent content for partially observed videos, they rely on expensive training data from video warping or 4D Gaussian field creation.
Moreover, they focus primarily on camera-controlled video generation~\cite{recammaster, cameractrlII} rather than 4D reconstruction, which does not guarantee consistency or efficiency for novel view playback like Gaussian Splatting.
In contrast, our method simply uses a pre-trained tracking model~\cite{cotracker} to generate masked training data and fine-tunes the model to provide additional supervision for 4D monocular reconstruction through iterative view expansion. Note that~\cite{viewcrafter, see3d} also employ diffusion models to inpaint warped videos, but they are limited to static scenes.

\section{Method}
\label{sec: method}

\begin{figure*}[t]
    \centering
    \includegraphics[width=\textwidth]{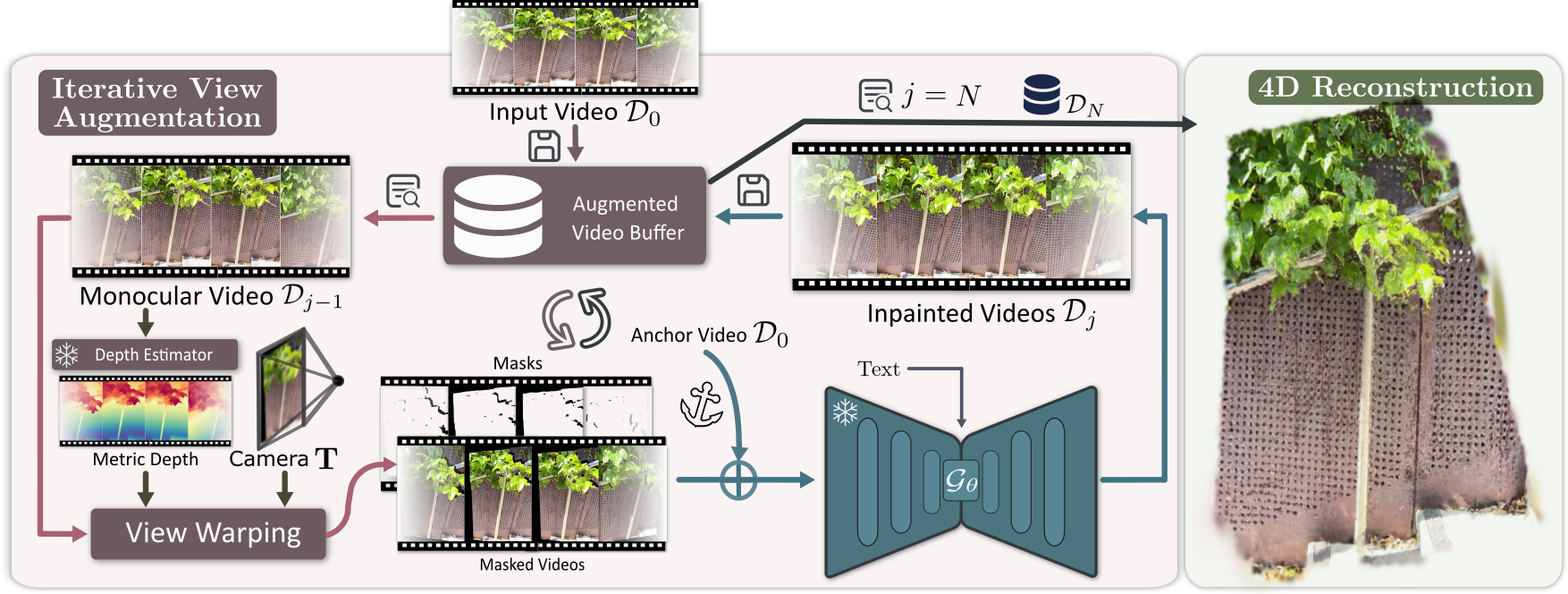}
    \vspace{-0.45cm}
    \caption{\textbf{4D reconstruction based on view augmentation.} Given an input monocular video, we first perform sparse reconstruction to obtain camera poses and align monocular depth to metric scale, forming an initial data buffer \(\mathcal{D}_0\). In each iterative view augmentation step, we select frames at each timestamp from the previous buffer \(\mathcal{D}_{j-1}\) and warp them to novel viewpoints using pre-defined camera poses \(\mathbf{T}\), creating new perspective videos with continuous invisible region masks. These masked videos, along with binary masks and an anchor video, are fed into our pre-trained anchor-conditioned video inpainting diffusion model to produce completed novel-view videos. We update the buffer \(\mathcal{D}_j\) with these enhanced videos, their metric depths and poses. Finally, both the original monocular video and all synthesized multi-view videos are used to supervise 4D scene reconstruction.}
   \label{fig: pipeline}
   \vspace{-0.2cm}
\end{figure*}

We present a novel approach for improving monocular 4D scene reconstruction through view augmentation using monocular depth priors and video diffusion models.
The key to enhancing the 4D monocular video synthesis is to warp the monocular video to augmented views based on the depth prior, followed by a video inpainting model to fill occluded regions. Our key components include a video inpainting model tailored for the task of view augmentation from a single monocular video~(\cref{sec: video inpainting diffusion model}), and an iterative augmented view generation strategy~(\cref{sec: diffusion-based view augmentation}). The original and the augmented videos are then used for 4D reconstruction~(\cref{sec: 4d reconstruction}). 
\cref{fig: pipeline} illustrates our 4D reconstruction based on view augmentation.

\subsection{Video Inpainting Diffusion Model}
\label{sec: video inpainting diffusion model}

We aim to learn a video inpainting diffusion model to inpaint the occluded or invisible areas in a masked video while ensuring spatial-temporal consistency. To align with our core objective of 4D reconstruction, it is essential to provide training data in which the regions requiring inpainting primarily result from camera motion.
Unlike previous methods, which require known camera poses~\cite{viewcrafter, cat4d}, or apply random mask to input videos~\cite{see3d}, our method leverages a pre-trained 2D tracking model~\cite{cotracker} for generating the training data for the video inpainting model, allowing us to leverage casual, unposed videos for training. 

\boldparagraph{Training Data Preparation} %
Given an unposed input video sequence \(\mathcal{V} = \{I_t\}_{t=1}^T\) with \(T\) frames, we first employ a pre-trained 2D tracking model~\cite{cotracker} to establish point correspondences across frames. Let \(\mathcal{P}_1 = \{p_1^i\}_{i=1}^N\) denote the set of \(N\) pixel points sampled from the first frame \(I_1\). The tracking model generates trajectories \(\mathcal{T} = \{(p_t^i)_{t=1}^T\}_{i=1}^N\) that record the positions of these initial points throughout the video sequence. For each subsequent frame \(I_t\) (\(t>1\)), we identify pixels that are not associated with any tracked point from \(\mathcal{P}_1\). These untracked pixels form our mask region \(\mathcal{M}_t\), as they likely correspond to areas that were previously occluded or newly revealed in the scene. Formally, we define the mask for frame \(t\) as:
\[
\mathcal{M}_t = \{q \in I_t \mid q \notin \{p_t^i\}_{i=1}^N\}
\]
where \(\{p_t^i\}_{i=1}^N\) represents the set of tracked points in frame \(t\). The masked frame can be obtained as:
\[
\mathcal{V}^m = \{I_t^m\}_{t=1}^T, \quad 
I_t^m = I_t \odot \mathcal{M}_t
\] 
where \(\odot\) indicates element-wise multiplication. This approach naturally captures regions that become visible due to object motion or camera movement, providing ideal training data for our video inpainting task.

\boldparagraph{Anchor-Conditioned Video Inpainting Diffusion Model} 
Given the masked video \(\mathcal{V}^m\) and corresponding binary mask \(\mathcal{M}\), 
we introduce a video diffusion model that inpaints occluded or invisible regions with spatial-temporal consistency while incorporating priors from unwarped videos. To leverage pre-trained video diffusion models effectively, we fine-tune an existing model~\cite{modelscope, videofusion} and adapt its architecture for video inpainting in view augmentation. Our modifications include processing the masked input and conditioning the model on unwarped videos to utilize full observed frames for context.

Specifically, we modify the standard video diffusion model in two ways. First, we extend the input processing to handle masked videos by encoding the incomplete video sequence through a VAE encoder, producing a \(4\)-channel latent representation \(\mathbf{z}_m \in \mathbb{R}^{T \times H \times W \times 4}\) that is concatenated with the corresponding resized binary mask \(\mathcal{M}' \in \mathbb{R}^{T \times H \times W \times 1}\)~\cite{sdinpaint} and the noisy latent \(\mathbf{z}_t \in \mathbb{R}^{T \times H \times W \times 4}\). Second, we introduce an anchor video mechanism, where frames sampled from the same video clip but at different timestamps serve as additional spatial-temporal context. Note that in our later 4D reconstruction stage, we use the original, unwarped monocular video as the anchor video. The anchor video is similarly encoded through the VAE into latent space \(\mathbf{z}_a \in \mathbb{R}^{T \times H \times W \times 4}\). During training, our modified U-Net receives a latent input \(\mathbf{z} = [\mathbf{z}_t; \mathbf{z}_m; \mathcal{M}'; \mathbf{z}_a]\), where these maps are concatenated along the channel dimension.
While \(\mathbf{z}_a\) are not spatially aligned with other maps using this simple concatenation, our 3D U-Net can effectively learn spatio-temporal correspondences through its hierarchical operations. The model is trained with the following objective:
\[\mathcal{L} = \mathbb{E}_{\mathbf{x}, t, \epsilon \sim \mathcal{N}(0,1)} \|\epsilon - \epsilon_\theta(\mathbf{z}_t, t, \mathbf{z}_m, \mathcal{M}', \mathbf{z}_a)\|_2^2\]
where \(\mathbf{z}_t = \alpha_t\mathbf{x} + \sigma_t\epsilon\) is the noised latent of the ground truth video \(\mathbf{x}\), and \(t\) is the diffusion timestep. 

\subsection{Diffusion-based View Augmentation}
\label{sec: diffusion-based view augmentation}

Given the trained video inpainting model, we augment the monocular input used for 4D reconstruction to generate multi-view supervision, as illustrated in \cref{fig: pipeline}.

\boldparagraph{Depth Estimation and Scale Alignment} 
Given a monocular video \(\mathcal{V} = \{I_t\}_{t=1}^T\) with \(T\) frames, we first utilize COLMAP~\cite{sfm, mvs} to obtain its camera intrinsics \(\mathbf{K}\), camera poses \(\{\mathbf{T}_t\}_{t=1}^T\), and a sparse point cloud \(\{X_i\}_{i=1}^M\), where each point \(X_i \in \mathbb{R}^3\) denotes the 3D coordinates reconstructed in the world coordinate system.

For depth estimation, we adopt a two-step approach that combines learning-based prediction with geometric refinement, as presented in~\cref{fig: pipeline}~(left). We first employ an off-the-shelf monocular depth estimation model (e.g.,~\cite{depthanythingv2}) to obtain initial depth maps \(\{\hat{D}_t\}_{t=1}^T\). Next, we align them with the metric scale using the sparse point cloud \(\{X_i\}_{i=1}^M\) from COLMAP via Least Square and RANSAC~\cite{ransac}, obtaining the metric depth maps \(\{D_t\}_{t=1}^T\). Please refer to supplementary~(~\cref{supp: depth_scale_alignment_and_view_warping})~for details of the scale alignment strategy.

\begin{figure}[t]
    \centering
    \includegraphics[width=\columnwidth]{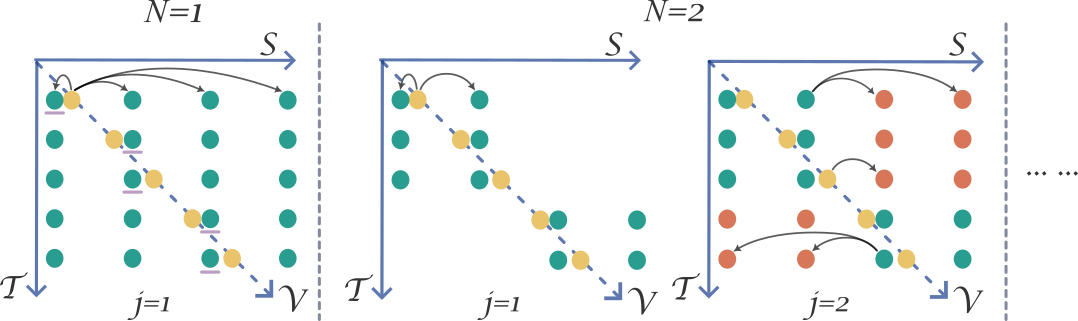}
    \vspace{-0.6cm}
    \caption{\textbf{Difference between Direct Warping and Iterative Warping.} \(\mathcal{S}\) and \(\mathcal{T}\) denote spatial and temporal dimensions, with \(\mathcal{V}\) being the input monocular video~(yellow dots indicating camera poses in 4D space). Left: With \(N=1\), we directly warp the input video to new perspectives~(green dots) in a single iteration. Warped frames are ranked by warping distance per timestamp and organized into a video (purple underline). Right: With \(N=2\), the video is first warped the closest pre-defined poses~(green dots, \(j=1\)), then additional perspectives~(red dots, \(j=2\)) are generated by selecting frames with minimal warping angles from existing frames. This iterative approach minimizes distortion and floaters caused by depth inaccuracies in large-angle warping, enhancing reconstruction quality.}
   \label{fig: view_augmentation}
   \vspace{-0.3cm}
\end{figure}

\boldparagraph{Iterative View Augmentation} 
Using the scale-aligned metric depth, we can warp the observed frames \(\mathcal{V} = \{I_t\}_{t=1}^T\) to unseen viewpoints for synthesizing multi-view observations, where each warped frame \(I_t' \in \mathbb{R}^{H\times W\times 3}\) is accompanied by a binary mask \(\mathcal{M}_t' \in \{0,1\}^{H\times W}\). However, due to inaccuracies in the depth priors~(e.g., bleeding edges),  directly warping the videos by a large angle introduces noticeable artifacts, which degrade the performance of 4D reconstruction. To address this, we propose an iterative view augmentation strategy, as illustrated in~\cref{fig: pipeline}~(middle), leveraging the fact that artifacts like bleeding edges are less severe when the warping angle is small. \cref{fig: view_augmentation} shows the difference between direct warping and iterative warping.

Specifically, we first pre-compute a set of target unseen camera viewpoints and perform iterative view augmentation over \(N\) iterations. As presented in \cref{fig: view_augmentation}, when \(N=1\)~(left), we directly warp the original video to all target frames in a single step. In contrast, when \(N>1\)~(right), we progressively expand the warped angles over multiple iterations. 
We start with a data buffer \(\mathcal{D}_0\) containing the input monocular video \(\mathcal{V}^0\), its aligned depths \(D^0\) and camera poses \(\mathbf{T}^0\), formally expressed as \(\mathcal{D}_0 = (\mathcal{V}^0, D^0, \mathbf{T}^0)\). At each iteration \(j\)~\((1\leq j \leq N)\), we select frames with the smallest warping angles from the data buffer \(\mathcal{D}_{j-1}\) at each timestamp \(t\) and warp them to target poses \(\mathbf{T}^{j}\), generating multiple new videos of \(T\) frames. The warped videos and corresponding masks, together with the anchor video, are then fed into our diffusion model to synthesize completed videos \(\hat{\mathcal{V}}^{j}\). Next, we estimate the metric depth \(D^{j}\) using our scale alignment strategy and update the data buffer as follows: \(\mathcal{D}_j = \mathcal{D}_{j-1} \cup (\hat{\mathcal{V}}^{j}, D^{j}, \mathbf{T}^{j})\). This process is repeated until all pre-defined target poses have been covered by each frame.
Furthermore, to avoid the inconsistency in 4D reconstruction supervisions caused by repeated inpainting across different iterations, we introduce a supervision mask \(S_t^{j} \in \{0,1\}^{H \times W}\) for each augmented frame, in which pixel value \(1\) indicates valid for supervision. Please refer to supplementary material~(~\cref{supp: supervision mask})~for more details.

\subsection{4D Reconstruction}
\label{sec: 4d reconstruction}

\boldparagraph{4D Scene Representation} 
Our method can effectively produce accurate, consistent, and high-fidelity novel view videos from monocular input, offering rich multi-view supervision for 4D scene reconstruction. It is compatible with various 4D reconstruction approaches, including those extending static representations with temporal dimensions~\cite{4dgs_time} and those modeling dynamic scenes via deformation or motion fields~\cite{4dgs_deformation, som}. In our experiments, we adopt a motion field-based 3DGS representation~\cite{som} to capture both geometric structure and temporal dynamics effectively.

\boldparagraph{Loss Function} 
We rely solely on the multi-view constraints from our synthesized novel views to guide reconstruction. To mitigate distortions from depth estimation errors and video diffusion artifacts, we propose to use \(\text{IV}\) RGB loss, a robust pixel-wise loss inspired by~\cite{iv-psnr}, as the supervisory signal for our augmented views. Specifically, we compute the L1 loss between each pixel in the rendered image and the closest \(3 \times 3\) pixels in the corresponding aug,mented supervision image, backpropagating only through the pixel with the minimum error. This loss, denoted as \(\mathcal{L}_{\text{IV}}\), improves reconstruction robustness by reducing sensitivity to slight misalignments. For the original frames, we use standard L1, SSIM and LPIPS loss to supervise the entire image. For our augmented frames, we use the loss function as follows:
\begin{small}
\[
    \mathcal{L} = \sum_{j=1}^N \sum_{t=1}^T \left( \lambda_r \mathcal{L}_{\text{IV}}^{t,j} + \lambda_s \mathcal{L}_{\text{ssim}}^{t,j} + \lambda_l \mathcal{L}_{\text{lpips}}^{t,j} \right)
\]
\end{small}
where each loss term is computed only in regions where the supervision mask \(\mathcal{S}_{t}^j\) equals 1. Specifically, for time step \(t\) in iteration \(j\):
\begin{align*}
    \mathcal{L}_{\text{IV}}^{t,j} &= \sum_{p} \mathcal{S}_{t}^j(p) \cdot \text{IV}(\overline{I}_t^j(p), \hat{I}_t^j(p)) \\
    \mathcal{L}_{\text{ssim}}^{t,j} &= \text{SSIM}(\overline{I}_t^j \odot \mathcal{S}_{t}^j, \hat{I}_t^j \odot \mathcal{S}_{t}^j) \\
    \mathcal{L}_{\text{lpips}}^{t,j} &= \text{LPIPS}(\overline{I}_t^j \odot \mathcal{S}_{t}^j, \hat{I}_t^j \odot \mathcal{S}_{t}^j)
\end{align*}
Here, \(p\) denotes pixel locations, \(\overline{I}_t^j\) and \(\hat{I}_t^j\) are the rendered and our augmented supervisions respectively, and \(\odot\) indicates element-wise multiplication. The hyperparameters \(\lambda_r\), \(\lambda_s\), and \(\lambda_l\) balance the contribution of each loss term.

\section{Experiments}
\label{sec: experiment}

\begin{figure*}[t]
    \centering
    \begin{tabular}{{@{}c@{\hspace{2pt}}c@{\hspace{2pt}}c@{\hspace{2pt}}c@{\hspace{2pt}}c@{}}}
        \includegraphics[width=0.19\textwidth]{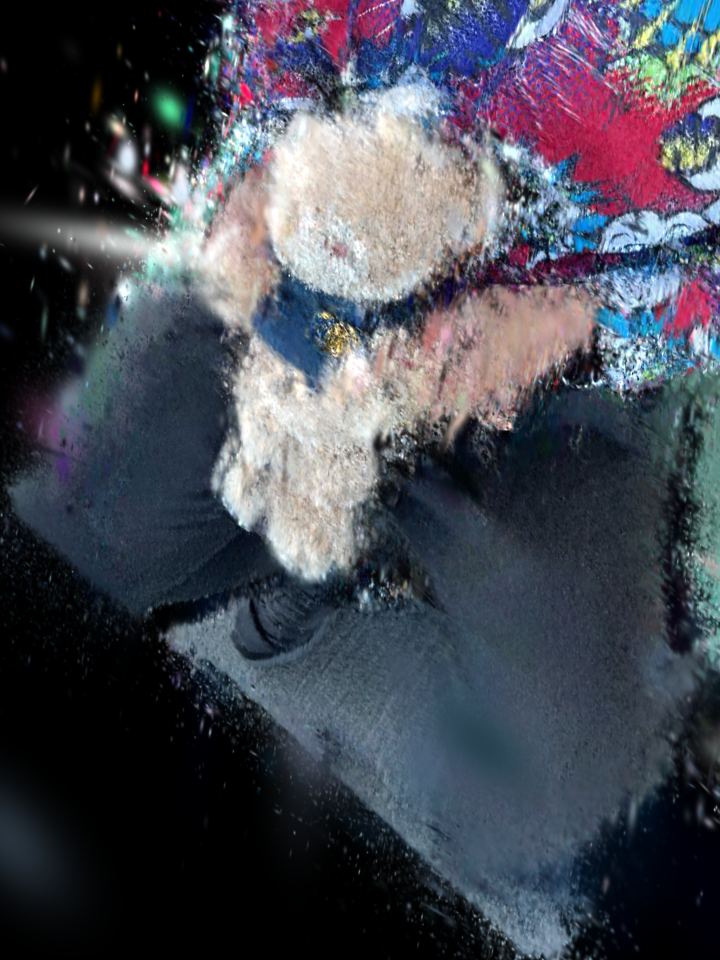} &
        \includegraphics[width=0.19\textwidth]{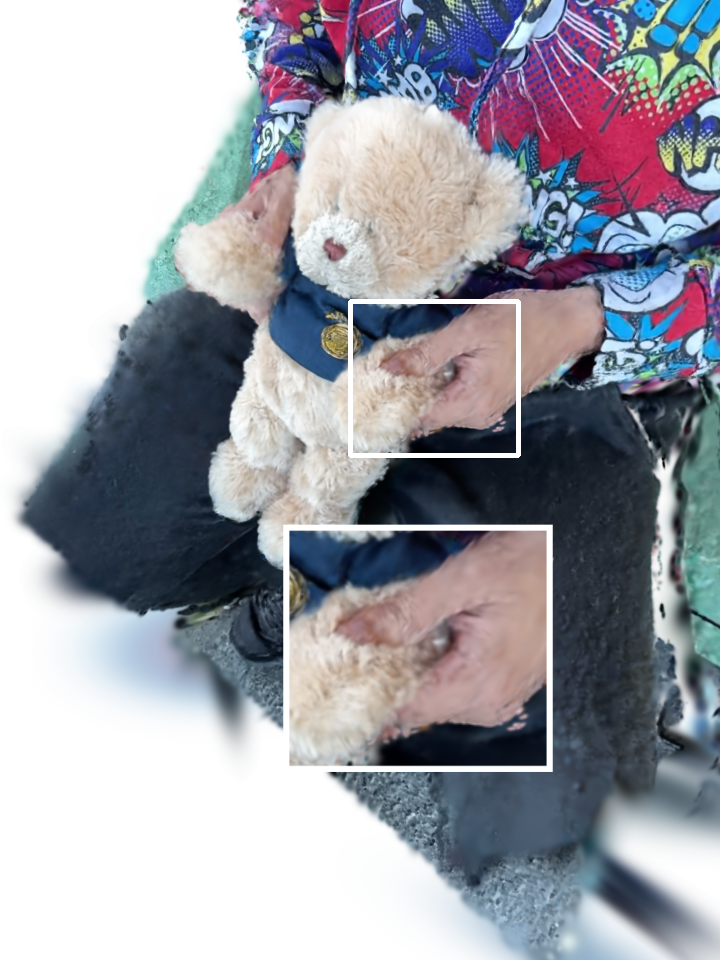} &
        \includegraphics[width=0.19\textwidth]{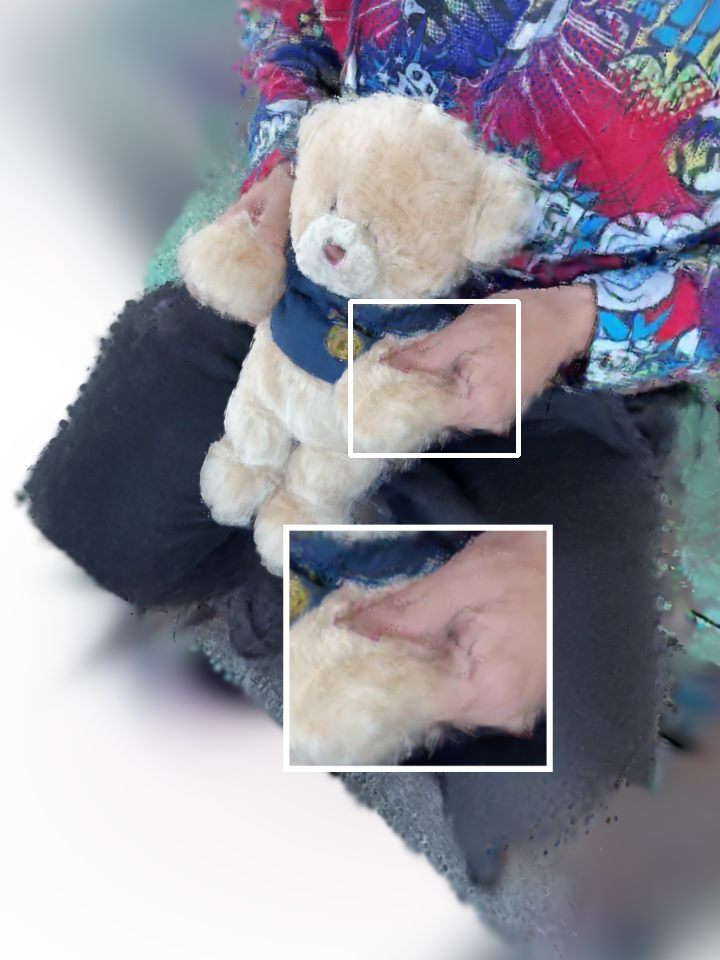} &
        \includegraphics[width=0.19\textwidth]{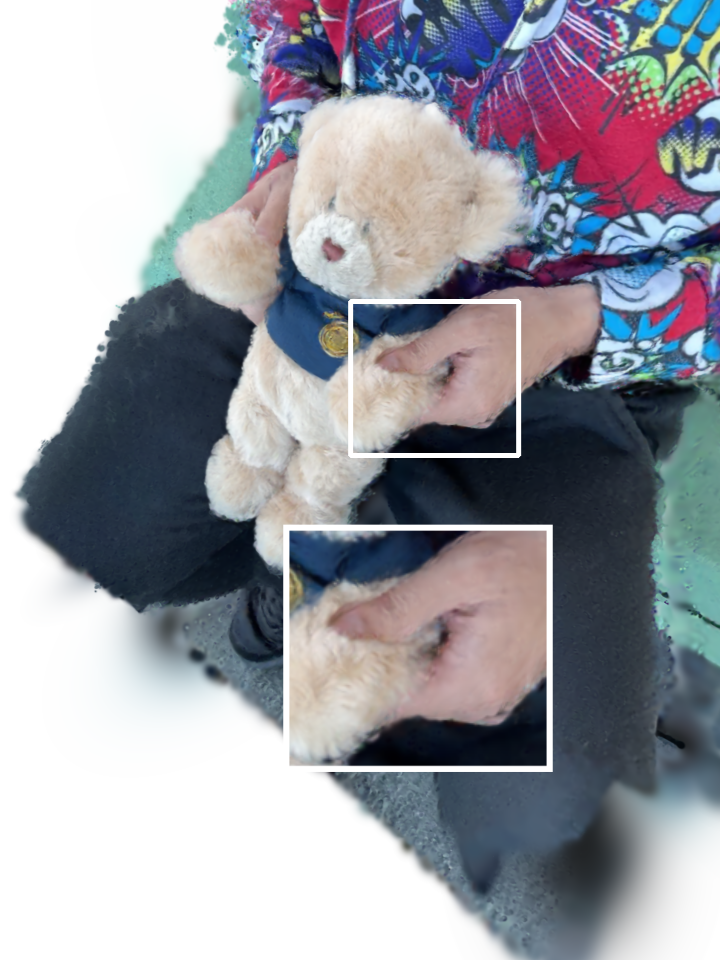} &
        \includegraphics[width=0.19\textwidth]{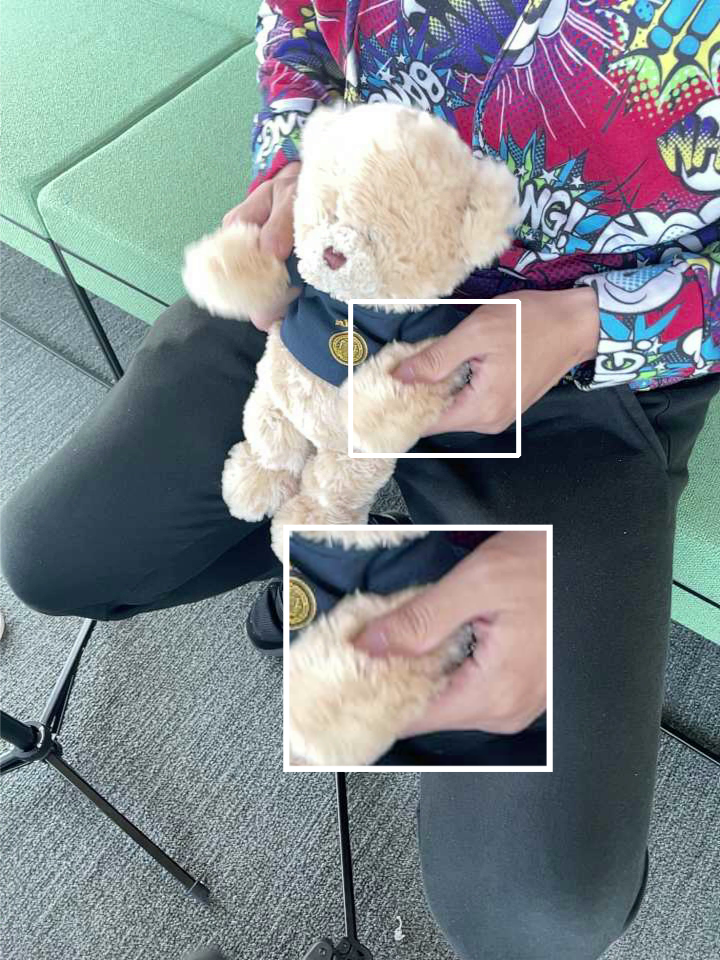} \\[-2.5pt]
        \includegraphics[width=0.19\textwidth]{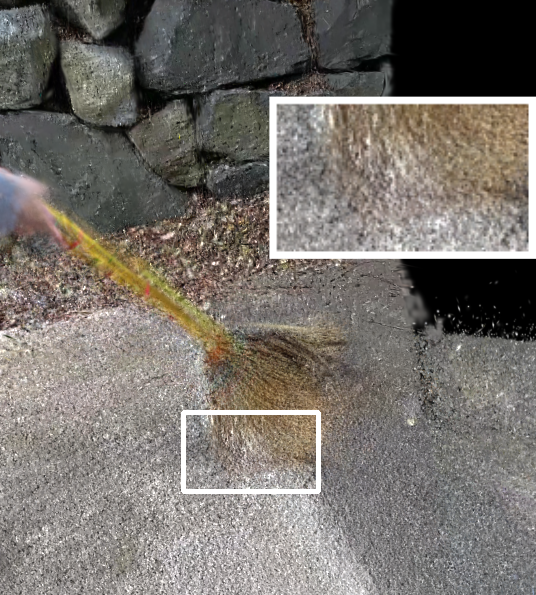} &
        \includegraphics[width=0.19\textwidth]{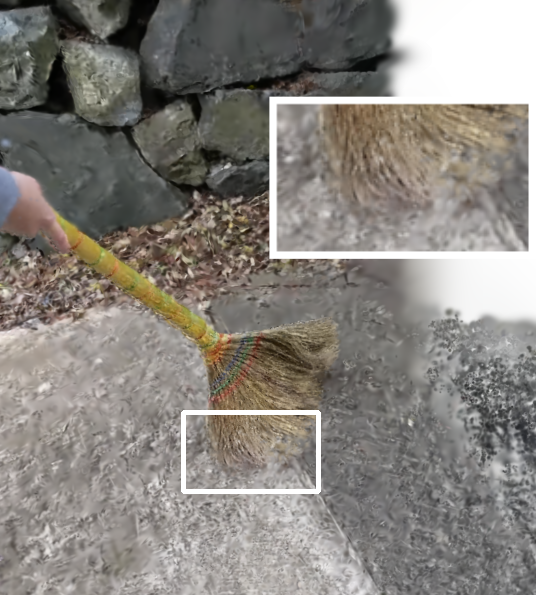} &
        \includegraphics[width=0.19\textwidth]{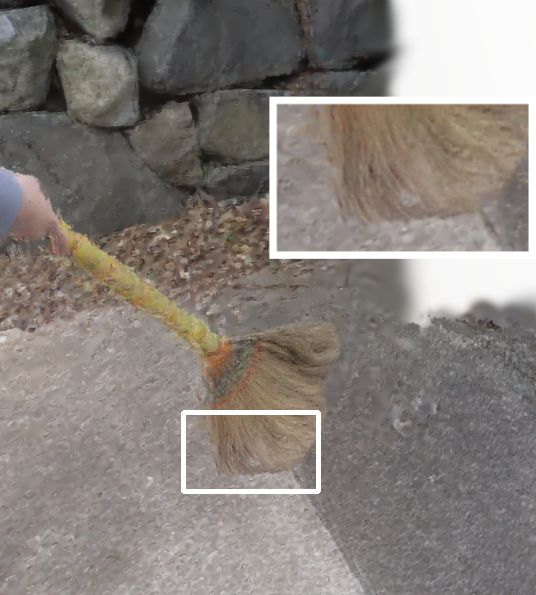} &
        \includegraphics[width=0.19\textwidth]{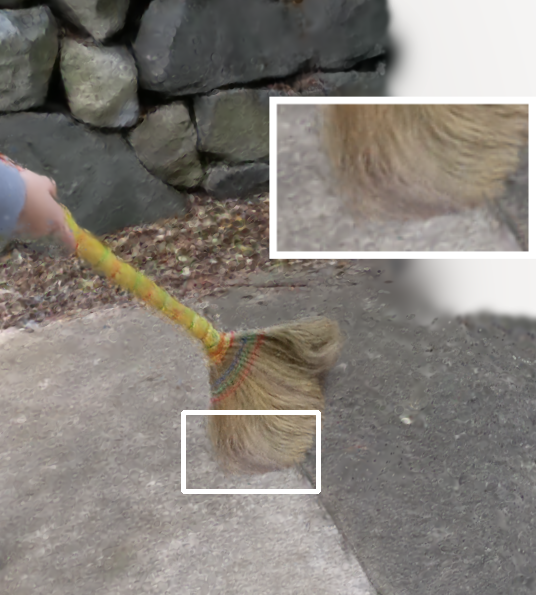} &
        \includegraphics[width=0.19\textwidth]{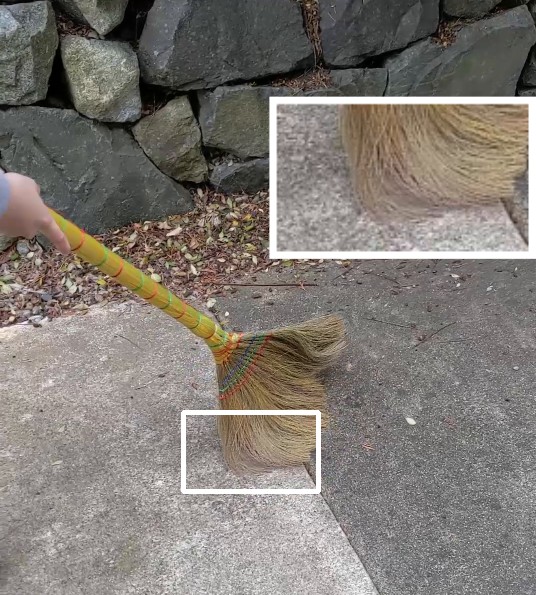} \\
        4D GS & Shape of Motion & CoCoCo & Ours & Ground Truth \\
    \end{tabular}
    \vspace{-0.25cm}
    \caption{\textbf{Qualitative comparison of dynamic scene reconstruction} on iPhone dataset and HyperNeRF dataset. The black holes in 4D GS and the white areas in Shape of Motion indicate regions where the input video lacks visibility. In contrast, our method effectively fills these invisible areas within the scenes, leveraging multi-view constraints and spatiotemporal priors to enhance reconstruction quality.}
    \label{fig: reconstruction_comparison}
    \vspace{-0.2cm}
\end{figure*}

In this section, we begin by comparing our method with state-of-the-art 4D scene reconstruction methods in~\cref{sec: scene reconstruction}. 
After that, we conduct ablation studies to evaluate our design choices in~\cref{sec: ablation study}. We highlight the \colorbox{lightred}{best}, \colorbox{lightorange}{second-best}, and \colorbox{lightyellow}{third-best} scores achieved on any metrics. Please refer to supplementary~(~\cref{supp: training_details})~for more training details.

\subsection{Monocular 4D Reconstruction}
\label{sec: scene reconstruction}

\boldparagraph{Dataset and Metrics} 
We evaluate dynamic scene reconstruction from monocular video using the iPhone~\cite{dycheck} and HyperNeRF~\cite{hypernerf} datasets. Our test sets comprise \(5\) iPhone and \(3\) HyperNeRF scenes, each with synchronized static cameras for novel-view synthesis evaluation. From the original video sequences, we select continuous clips with a few dozen frames, chosen for their large perspective differences from the test view to emphasize our method's efficacy. We assess reconstruction quality using mPSNR, mSSIM, and mLPIPS~\cite{dycheck} for dynamic foreground, static background, and the overall scene, employing a co-visible mask to specify visible areas for reconstruction.

\boldparagraph{Comparison Baselines} 
We compare our method against 4D GS~\cite{4dgs_time}, Shape of Motion~\cite{som}, CoCoCo~\cite{cococo}, StereoCrafter~\cite{stereocrafter}, and ViewCrafter~\cite{viewcrafter}. CoCoCo and StereoCrafter are designed for video inpainting, while ViewCrafter focuses on 3D-aware inpainting. We integrate their diffusion models into our pipeline for warped video inpainting to validate the necessity of our anchor-conditioned model. For fairness, we provide CoCoCo with the same text prompt for each scene and use the first frame of our anchor video as the image conditioning input for ViewCrafter. Additionally, we compare against the state-of-the-art monocular reconstruction method Shape of Motion~\cite{som} and the general 4D reconstruction approach 4D GS~\cite{4dgs_time}.

\begin{table}[t]
    \centering
    \small\relsize{0.5}
    \begin{tabular}{@{}l@{\hspace{23pt}}c@{\hspace{6pt}}c@{\hspace{6pt}}c@{\hspace{6pt}}c}
        \toprule
        Condition & PSNR~$\uparrow$ & SSIM~$\uparrow$ & LPIPS~$\downarrow$ & FVD~$\downarrow$ \\
        \midrule
        w/o anchor & 25.34 & 0.8053 & 0.1056 & 18.99 \\
        w anchor~(ours) & \bf\best{27.22} & \bf\best{0.8223} & \bf\best{0.0801} & \bf\best{14.30} \\
        \bottomrule
    \end{tabular}
    \vspace{-0.25cm}
    \caption{\textbf{Quantitative ablation study on anchor video condition} on our processed \(5K\) videos from OpenVid-1M.}
    \label{tab: ablation_anchor}
    \vspace{-0.2cm}
\end{table}

\begin{table*}[htbp]
    \centering
    \footnotesize\relsize{0.5}
    \begin{adjustbox}{max width=\linewidth}
    \begin{tabular}{lccccccccc}
    \toprule
    \multicolumn{1}{l}{\textbf{Dataset}} & \multicolumn{3}{c}{Dynamic Part} & \multicolumn{3}{c}{Static Part} & \multicolumn{3}{c}{All} \\
    \cmidrule(lr){1-1} \cmidrule(lr){2-4} \cmidrule(lr){5-7} \cmidrule(lr){8-10}
    Method & mPSNR~$\uparrow$ & mSSIM~$\uparrow$ & mLPIPS~$\downarrow$ & mPSNR~$\uparrow$ & mSSIM~$\uparrow$ & mLPIPS~$\downarrow$ & mPSNR~$\uparrow$ & mSSIM~$\uparrow$ & mLPIPS~$\downarrow$ \\
    \midrule
    \multicolumn{10}{l}{\textbf{iPhone Dataset}} \\
    \midrule
    4D GS & \sbest{12.28} & 0.8757 & 0.5249 & 14.54 & 0.4870 & 0.5594 & 14.01 & 0.3877 & 0.5939 \\
    Shape of Motion & 11.87 & \tbest{0.8894} & \tbest{0.4778} & 15.42 & 0.5744 & \sbest{0.5082} & 14.56 & \tbest{0.4570} & \tbest{0.5292} \\
    CoCoCo & \tbest{12.20} & 0.8964 & \sbest{0.4650} & \sbest{15.85} & 0.5809 & \tbest{0.5130} & \sbest{14.99} & 0.4701 & \sbest{0.5280} \\
    StereoCrafter & 12.09 & \sbest{0.8989} & 0.4811 & 15.60 & \sbest{0.5989} & 0.5593 & 14.85 & \sbest{0.4945} & 0.5676 \\
    ViewCrafter & 12.13 & \tbest{0.8985} & 0.5053 & \tbest{15.82} & \tbest{0.5946} & 0.5315 & \tbest{14.94} & \tbest{0.4888} & 0.5772 \\
    Ours & \bf\best{12.34} & \bf\best{0.8990} & \bf\best{0.4516} & \bf\best{16.31} & \bf\best{0.6088} & \bf\best{0.4726} & \bf\best{15.20} & \bf\best{0.5004} & \bf\best{0.4930} \\
    \midrule
    \multicolumn{10}{l}{\textbf{HyperNeRF Dataset}} \\
    \midrule
    4D GS & 17.93 & 0.9556 & 0.3819 & 18.32 & 0.4737 & \sbest{0.4409} & 18.24 & 0.4200 & \sbest{0.4450} \\
    Shape of Motion & 17.70 & 0.9609 & \sbest{0.3027} & \sbest{19.12} & 0.5659 & \tbest{0.4520} & 18.82 & 0.5176 & \tbest{0.4589} \\
    CoCoCo & 18.47 & 0.9648 & \tbest{0.3048} & \tbest{19.11} & 0.5599 & 0.4627 & \sbest{19.00} & 0.5147 & 0.4692 \\
    StereoCrafter & \tbest{18.51} & \sbest{0.9660} & 0.3343 & 18.98 & \tbest{0.5677} & 0.5171 & 18.86 & \tbest{0.5231} & 0.5181 \\
    ViewCrafter & \sbest{18.53} & \tbest{0.9658} & 0.3228 & 18.98 & \sbest{0.5691} & 0.4860 & \tbest{18.91} & \sbest{0.5259} & 0.4888 \\
    Ours & \bf\best{18.85} & \bf\best{0.9666} & \bf\best{0.3020} & \bf\best{19.57} & \bf\best{0.5910} & \bf\best{0.4407} & \bf\best{19.45} & \bf\best{0.5446} & \bf\best{0.4449} \\
    \bottomrule
    \end{tabular}
    \end{adjustbox}
    \vspace{-0.25cm}
    \caption{\textbf{Quantitative comparison of dynamic scene reconstruction} on iPhone dataset and HyperNeRF dataset.}
    \label{tab: reconstruction_comparison}
    \vspace{-0.2cm}
\end{table*}

\boldparagraph{Results} 
The results are shown in~\cref{fig: reconstruction_comparison} and~\cref{tab: reconstruction_comparison}. It can be observed that our pipeline significantly improves reconstruction quality over 4D GS and Shape of Motion, which exhibit outliers or over-sharpening artifacts. In contrast, our method not only produces a smoother results but also better preserves dynamic foreground details. Additionally, it fills missing regions invisible in the input video, addressing the inner holes seen in 4D GS and Shape of Motion. Compared to CoCoCo, which also achieves high-quality reconstruction, our method offers greater overall clarity, finer details—especially in dynamic areas—and more contextually consistent inpainting, thanks to our anchor-based video inpainting model. Quantitatively, our approach outperforms all baselines across all metrics, further demonstrating its effectiveness in 4D reconstruction.

\subsection{Ablation Study}
\label{sec: ablation study} 

We ablate various components of our method on the iPhone dataset and HyperNeRF dataset for 4D reconstruction to verify design choices of our approach.

\begin{figure}[t]
    \centering
    \small
    \begin{tabular}{{@{}c@{\hspace{2pt}}c@{\hspace{2pt}}c@{}}}
        \includegraphics[width=0.15\textwidth]{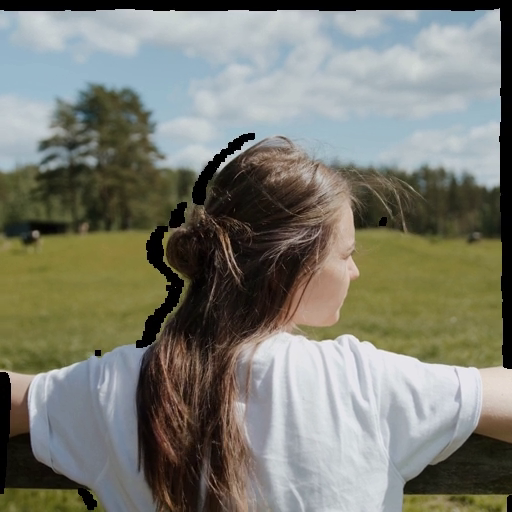} &
        \includegraphics[width=0.15\textwidth]{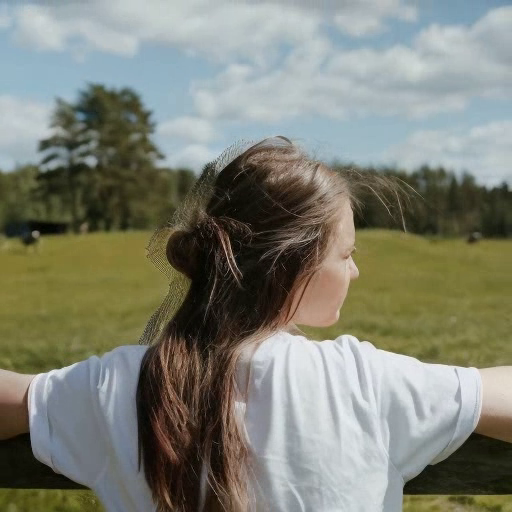} &
        \includegraphics[width=0.15\textwidth]{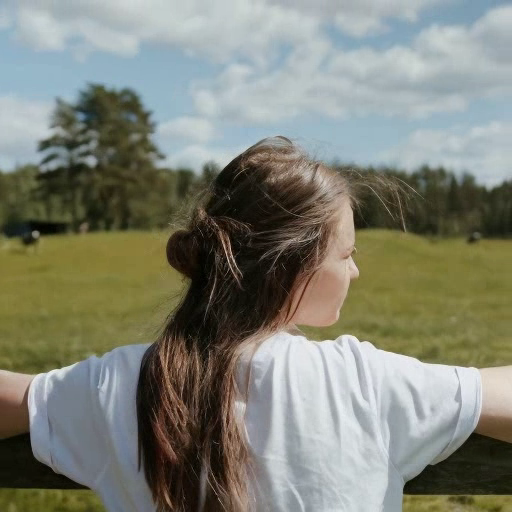}\\[-2.5pt]
        \includegraphics[width=0.15\textwidth]{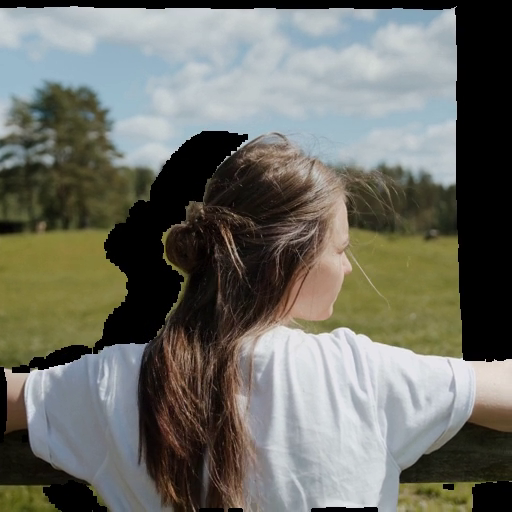} &
        \includegraphics[width=0.15\textwidth]{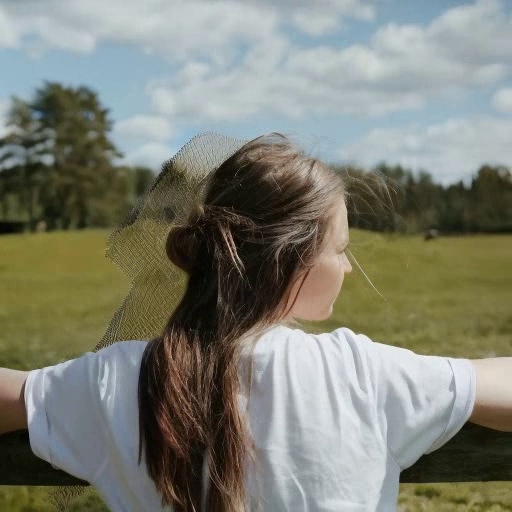} &
        \includegraphics[width=0.15\textwidth]{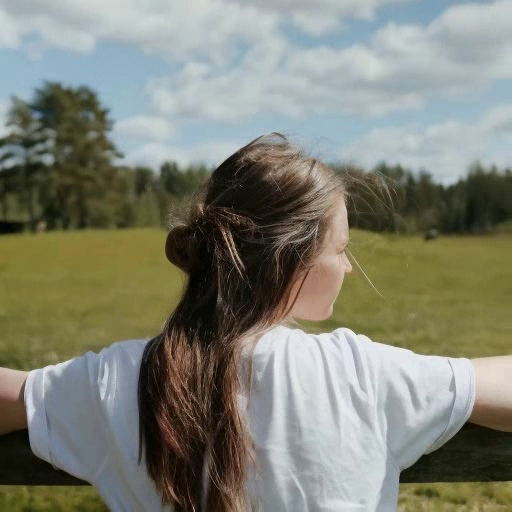}\\
        Masked Sequence & w/o anchor &  w anchor~(ours) \\
    \end{tabular}
    \vspace{-0.25cm}
    \caption{\textbf{Quantitative ablation study on anchor video condition} on our processed \(5K\) videos from OpenVid-1M.}
    \label{fig: ablation_anchor}
    \vspace{-0.4cm}
\end{figure}

\boldparagraph{Effect of Anchor condition} 
We explored the impact of anchor video condition on the video inpaint results of our diffusion model, as presented in~\cref{tab: ablation_anchor} and~\cref{fig: ablation_anchor}. Without the anchor condition, the model's filling capability weakens, resulting in noticeable artifacts. In contrast, after incorporating the anchor video, the filled areas become more consistent with the surrounding content, significantly reducing artifacts. This demonstrates that our model effectively leverages spatiotemporal priors from the anchor video.

\begin{figure}[t]
    \centering
    \footnotesize
    \setlength{\tabcolsep}{0.5pt}
    \vspace{-0.15cm}
    \begin{tabular}{@{}cccc@{}}
        \includegraphics[width=0.245\columnwidth]{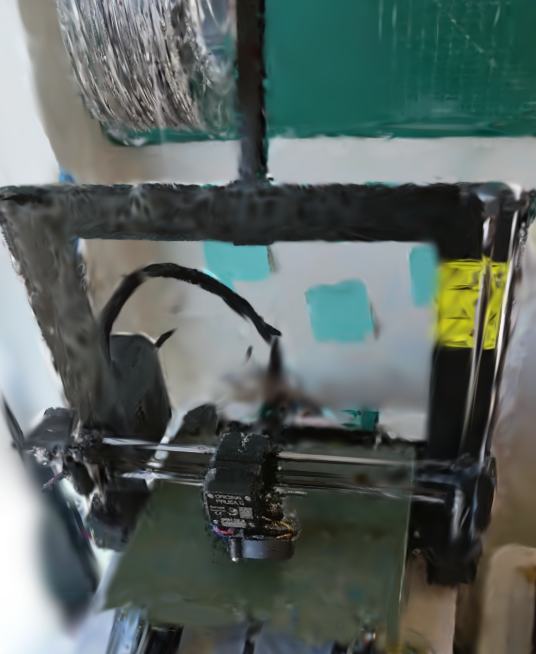} &
        \includegraphics[width=0.245\columnwidth]{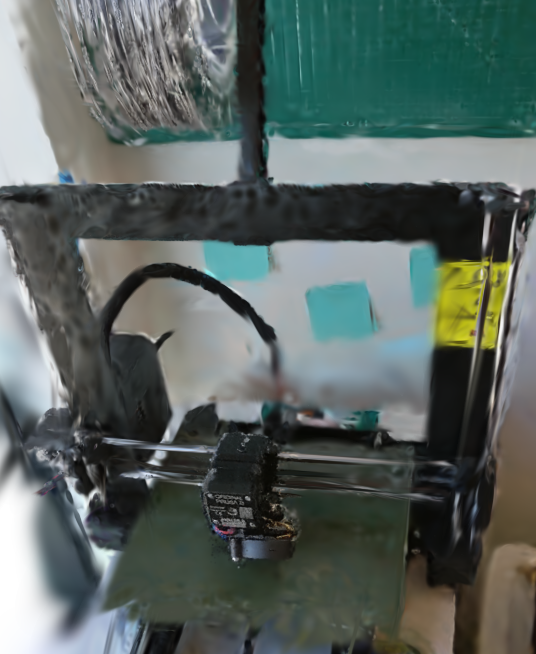} &
        \includegraphics[width=0.245\columnwidth]{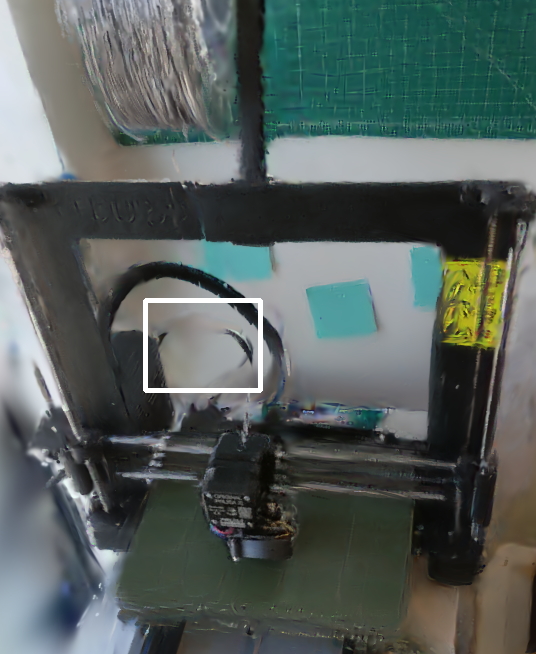} &
        \includegraphics[width=0.245\columnwidth]{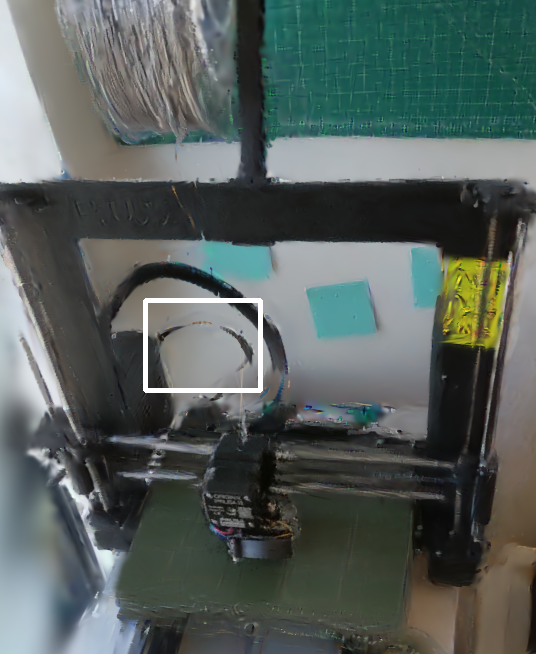} \\[-2.0pt]
        \includegraphics[width=0.245\columnwidth]{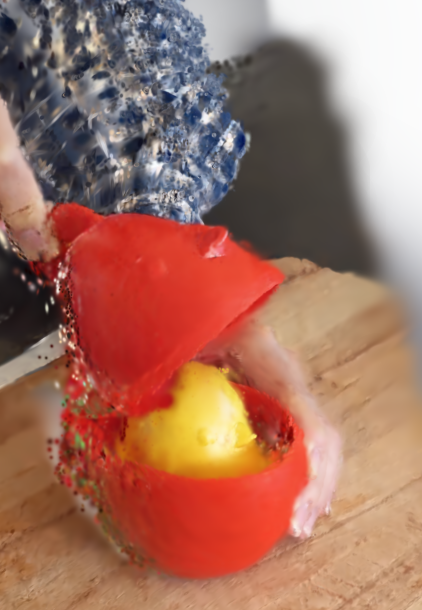} &
        \includegraphics[width=0.245\columnwidth]{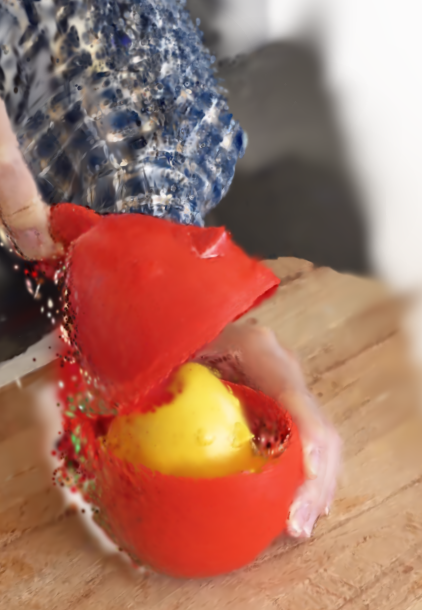} &
        \includegraphics[width=0.245\columnwidth]{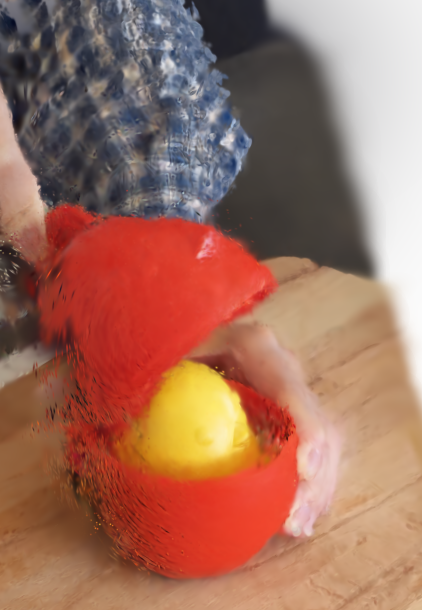} &
        \includegraphics[width=0.245\columnwidth]{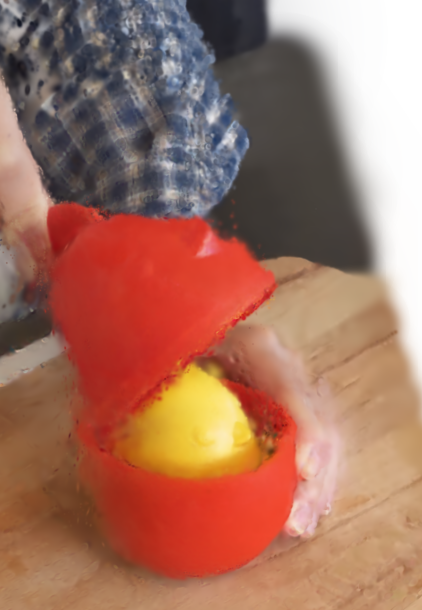}\\
        (a) & (b) & (c) & (d) \\
    \end{tabular}
    \vspace{-0.25cm}
    \caption{\textbf{Qualitative ablation study on view warping and video inpainting} on HyperNeRF dataset. (a) w/o warping, w/o inpainting, w/o depth; (b) w/o warping, w/o inpainting, w/ depth; (c) w/ warping, w/o inpainting; (d) w/ warping, w/ inpainting (ours). }
    \label{fig: ablation_warp_inpainting}
    \vspace{-0.4cm}
\end{figure}

\boldparagraph{Effect of View Warping and Video inpainting} 
To assess the effectiveness of our view augmentation, which includes view warping and video inpainting, and to compare it with direct depth supervision loss $\cL_{D}$,  we compare our full method against: (a) no view augmentation or depth supervision, (b) direct depth supervision without view augmentation, and (c) iterative view warping without video inpainting.
For fairness, all experiments use the same Gaussian initialization, and our view augmentation is set to \(6\) iterations. The results are presented in~\cref{tab: ablation_warp_inpainting} and~\cref{fig: ablation_warp_inpainting}. 

It can be observed that directly leveraging the depth for supervision~(b)~only leads to minor improvement compared with~(a). In contrast, using the same depth to synthesize augmented training views leads to larger improvement even without inpainting~(c). This verifies that directly adopting geometric priors for supervision may not linearly correlate with the reconstruction performance in the RGB space.
Moreover, with our inpainting model to complete the warped views (d), reconstruction further improves, correctly recovering previously invisible areas and reducing flying artifacts. This demonstrates that leveraging depth priors via warping benefits reconstruction, while integrating diffusion priors further enhances completion and quality in 4D reconstruction.

\begin{table}[t]
    \centering
    \footnotesize
    \begin{tabular}{@{}l@{\hspace{7pt}}c@{\hspace{4pt}}c@{\hspace{4pt}}c}
        \toprule
        Method & mPSNR~$\uparrow$ & mSSIM~$\uparrow$ & mLPIPS~$\downarrow$ \\
        \midrule
        (a) w/o warp, w/o inpaint, w/o $\cL_{D}$ & 16.04 & 0.4617 & 0.5249 \\
        (b) w/o warp, w/o inpaint, w/ $\cL_{D}$ & \tbest{16.12} & \tbest{0.4699} & \tbest{0.5084} \\
        (c) w/ warp, w/o inpaint, w/o $\cL_{D}$ & \sbest{16.44} & \sbest{0.4913} & \sbest{0.4876} \\
        (d) w/ warp, w/ inpaint, w/o $\cL_{D}$~(ours) & \bf\best{16.80} & \bf\best{0.5170} & \bf\best{0.4750} \\
        \bottomrule
    \end{tabular}
    \vspace{-0.25cm}
    \caption{\textbf{Quantitative ablation study on view warping and video inpainting} on iPhone dataset and HyperNeRF dataset.}
    \label{tab: ablation_warp_inpainting}
    \vspace{-0.12cm}
\end{table}

\begin{table}[t]
    \centering
    \small\relsize{-0.49}
    \begin{tabular}{@{}lc@{\hspace{9pt}}c@{\hspace{9pt}}c}
        \toprule
        Iterations & mPSNR~$\uparrow$ & mSSIM~$\uparrow$ & mLPIPS~$\downarrow$ \\
        \midrule
        w/o view aug., w/o depth & 16.04 & 0.4617 & 0.5249 \\
        \(N=1, j=1\) & 15.80 & 0.4711 & \tbest{0.4959} \\
        \(N=6, j=2\) & \tbest{16.41} & \tbest{0.5054} & 0.4964 \\
        \(N=6, j=4\) & \sbest{16.56} & \sbest{0.5104} & \sbest{0.4868} \\
        \(N=6, j=6\)~(ours) & \bf\best{16.80} & \bf\best{0.5170} & \bf\best{0.4750} \\
        \bottomrule
    \end{tabular}
    \vspace{-0.25cm}
    \caption{\textbf{Quantitative ablation study on iterations} on iPhone dataset and HyperNeRF dataset.}
    \label{tab: ablation_iteration}
    \vspace{-0.15cm}
\end{table}

\begin{figure}[t]
    \centering
    \footnotesize
    \begin{tabular}{@{}c@{\hspace{-0.5pt}}c@{\hspace{-0.5pt}}c@{\hspace{-0.5pt}}c@{\hspace{-0.5pt}}c@{}}
        \includegraphics[width=0.20\columnwidth]{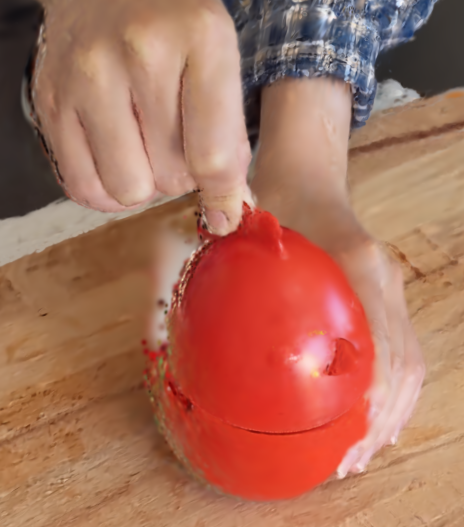} &
        \includegraphics[width=0.20\columnwidth]{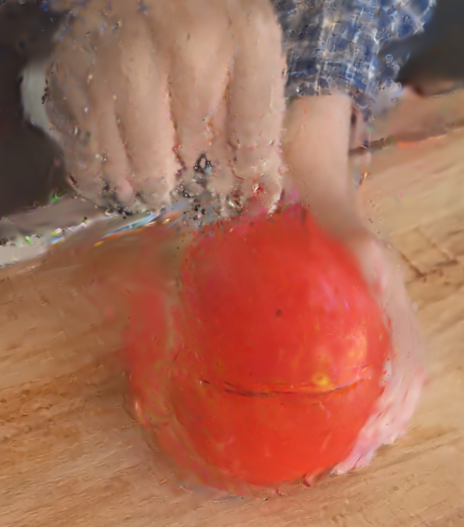} &
        \includegraphics[width=0.20\columnwidth]{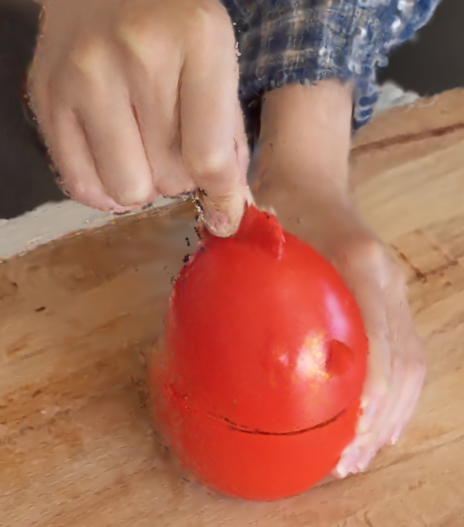} &
        \includegraphics[width=0.20\columnwidth]{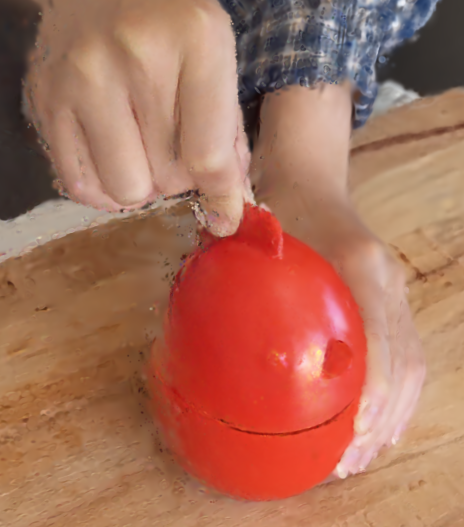} &
        \includegraphics[width=0.20\columnwidth]{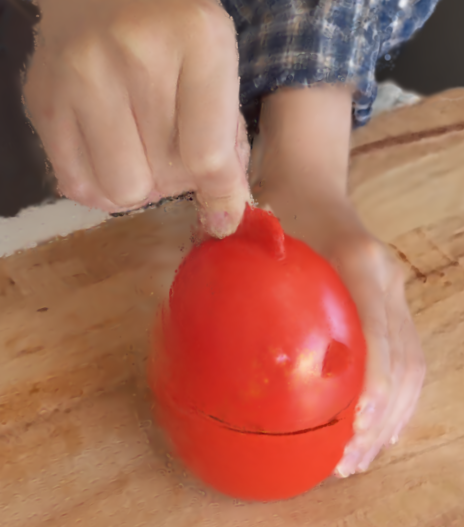} \\
        w/o view aug. & 
        \begin{tabular}[t]{c}\(N=1\), \\[-1pt] \(j=1\)\end{tabular} & 
        \begin{tabular}[t]{c}\(N=6\), \\[-1pt] \(j=2\)\end{tabular} & 
        \begin{tabular}[t]{c}\(N=6\), \\[-1pt] \(j=4\)\end{tabular} & 
        \begin{tabular}[t]{c}\(N=6\), \\[-1pt] \(j=6\)\end{tabular} \\
    \end{tabular}
    \vspace{-0.3cm}
    \caption{\textbf{Qualitative ablation study on iterations} on HyperNeRF dataset.}
    \label{fig: ablation_iteration}
    \vspace{-0.25cm}
\end{figure}

\boldparagraph{Effect of Iterations} 
We investigated the impact of iteration count on reconstruction results, as shown in~\cref{tab: ablation_iteration} and~\cref{fig: ablation_iteration}. When no iterations are performed (w/o view aug), meaning only a single monocular video is used for supervision, the rendered image exhibits holes and noticeable artifacts at object edges. After applying view augmentation, we obtain more comprehensive observations of the scene. However, directly warping the initial video into multiple views at once~(\(N=1, j=1\)) results in poor reconstruction due to large-angle warping, which introduces severe artifacts when transforming to distant viewpoints~(e.g., bleeding artifacts are more noticeable at larger viewing angles). In contrast, our iterative strategy gradually enhances viewpoints, reducing warping angles and minimizing the impact of depth inaccuracies. Even with a small number of iterations~(e.g., \(2\) or \(4\)), the scene is better filled, and multiview supervision helps mitigate floaters. As iterations increase, more comprehensive views of the scene are obtained for each timestamp, leading to more robust reconstruction. Although the overall scene coverage remains the same for \(N=1, j=1\) and \(N=6, j=6\), additional iterations reduce warping distortion and improve reconstruction fidelity. Quantitative metrics further confirm that increasing the number of iterations leads to progressively better reconstruction results.

\boldparagraph{Effect of Depth Estimation Methods} 
Next, we evaluate various depth estimation methods~\cite{depthanythingv2, depth_pro, depthcrafter, videodepthanything} to assess their impact, as shown in~\cref{tab: ablation_depth_estimation}. We observe that different methods yield similar results, indicating our method's robustness against monocular depth estimation variations. This may be attributed to aligning monocular depth predictions with COLMAP’s scale, along with our interactive view warping and robust RGB loss. By default, we use Depth Anything V2~\cite{depthanythingv2}, as it achieves slightly better performance than the others.
\begin{table}[t]
    \centering
    \footnotesize
    \resizebox{\linewidth}{!}{
    \begin{tabular}{@{}lccc}
        \toprule
        Model & mPSNR~$\uparrow$ & mSSIM~$\uparrow$ & mLPIPS~$\downarrow$ \\
        \midrule
        DepthCrafter & 16.76 & \tbest{0.5135} & \tbest{0.4835} \\
        Depth Pro & \sbest{16.79} & 0.5121 & \sbest{0.4807} \\
        Video Depth Anything & \sbest{16.79} & \sbest{0.5159} & 0.4875 \\
        Depth Anything V2~(ours) & \bf\best{16.80} & \bf\best{0.5170} & \bf\best{0.4750} \\
        \bottomrule
    \end{tabular}}
    \vspace{-0.25cm}
    \caption{\textbf{Quantitative ablation study on various depth estimation models} on iPhone dataset.}
    \label{tab: ablation_depth_estimation}
\end{table}

\boldparagraph{Effect of Loss Function} 
Lastly, we also evaluated the effectiveness of \(\text{IV}\) RGB loss on the robustness of the results. Instead of using \(\mathcal{L}_\text{IV}\) for RGB supervision, we directly applied the L1 loss. The results, shown in~\cref{tab: ablation_loss} and~\cref{fig: ablation_loss} (best viewed zoomed in), reveal that L1 loss leads to blurrier renderings due to misaligned supervision. In contrast, our method backpropagates errors only to the pixel with the minimum difference within adjacent patches, effectively reducing blurring caused by distortion.

\newcommand{\bestloss}[1]{%
  \begin{tabular}{>{\columncolor{lightred}[8.2pt][6.5pt]}c<{}}#1\end{tabular}%
}
\newcommand{\sbestloss}[1]{%
  \begin{tabular}{>{\columncolor{lightorange}[0pt][0pt]}c<{}}#1\end{tabular}%
}
\newcommand{\tbestloss}[1]{%
  \begin{tabular}{>{\columncolor{lightyellow}[0pt][0pt]}c<{}}#1\end{tabular}%
}

\begin{table}[t]
    \centering
    \begin{tabular}[c]{@{}ll@{}}
        \begin{minipage}[c]{0.4\columnwidth}
            \centering
            \footnotesize
            \vspace{-0.15cm}
            \begin{tabular}{{@{}c@{\hspace{0.5pt}}c@{}}}
                L1  & IV~(ours) \\
                \includegraphics[width=0.5\columnwidth]{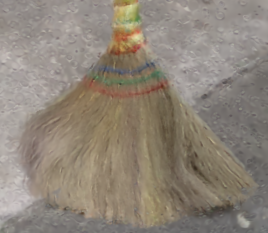} &
                \includegraphics[width=0.5\columnwidth]{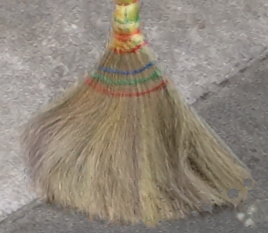}
            \end{tabular}
            \vspace{-0.3cm}
            \captionof{figure}{\textbf{Qualitative ablation on loss function} on HyperNeRF dataset.}
            \label{fig: ablation_loss}
        \end{minipage}
        &
        \begin{minipage}[c]{0.55\columnwidth}
            \centering
            \scriptsize
            \vspace{0.12cm}
            \begin{tabular}{@{}l@{\hspace{4pt}}c@{\hspace{2pt}}c@{\hspace{2pt}}c@{}}
                \toprule
                Loss & mPSNR~$\uparrow$ & mSSIM~$\uparrow$ & mLPIPS~$\downarrow$ \\
                \midrule
                L1 & 16.50 & 0.5118 & 0.5097 \\
                IV~(ours) & \bf\bestloss{16.80} & \bf\bestloss{0.5170} & \bf\bestloss{0.4750} \\
                \bottomrule
            \end{tabular}
            \vspace{-0.03cm}
            \captionof{table}{\textbf{Quantitative ablation study on loss function} on iPhone dataset and HyperNeRF dataset.}
            \label{tab: ablation_loss}
        \end{minipage}
    \end{tabular}
    \vspace{-0.4cm}
\end{table}

\section{Conclusion}
\label{sec: conclusion}

We introduced~\method{}, a novel approach for 4D monocular reconstruction that combines geometric priors with video diffusion models. By reformulating view augmentation as a video inpainting task, our method leverages unposed web videos for training and employs an iterative view expansion strategy with a robust loss to minimize artifacts from depth errors. Experiments demonstrate that~\method{} consistently outperforms existing methods across dynamic scenes, producing more coherent and detailed novel view synthesis results while effectively filling unobserved regions. Our approach bridges geometric-based methods and generative models, enabling high-quality 4D reconstruction from casual monocular videos, with applications in virtual reality and content creation. Future work will focus on enabling plausible scene extrapolation beyond observed regions, and improving robustness to complex camera motions and scene dynamics.

{
    \small
    \bibliographystyle{ieeenat_fullname}
    \bibliography{main}
}

\clearpage
\setcounter{page}{1}
\maketitlesupplementary

\section{Implemenentation Details}

\subsection{Training Data Curation}
\label{supp: data_curation}

Based on our proposed pipeline that utilizes pre-trained 2D tracking model to generate masked videos, we curated our training dataset from OpenVid-1M~\cite{openvid-1m}. To ensure the quality and effectiveness of our training data, we applied two filtering criteria. First, we removed videos with fixed viewpoints using the camera motion annotations provided by OpenVid-1M, as they do not offer the viewpoint variation needed for robust view synthesis and 3D understanding. Second, we filtered out sequences with insufficient dynamic content, despite having camera motion. Specifically, we computed the ratio of masked regions to the total video area:
\[
R_v = \frac{1}{T}\sum_{t=1}^T \frac{|\mathcal{M}_t|}{|I_t|}
\]
where \(|\mathcal{M}_t|\) represents the number of pixels in the mask region of frame \(t\), and \(|I_t|\) is the total number of pixels in frame \(t\). Videos with a ratio \(R_v\) below a threshold \(\tau\)~(\(\tau=0.98\)) were excluded from our training set, as they indicate limited scene dynamics and thus provide insufficient novel content for learning inpainting patterns.

Through this filtering process, our final dataset comprises \(150K\) video clips totaling \(7.5M\) frames, each containing rich viewpoint variations and dynamic scene content. The simplicity and automation of our curation pipeline makes it particularly scalable, enabling continuous dataset expansion as new video content becomes available online.

\subsection{Training Details of Diffusion Model}
\label{supp: training_details}

We fine-tuned our video inpainting diffusion model based on a text-to-video diffusion model~\cite{modelscope, videofusion, svd}. The training data is selected from OpenVid-1M~\cite{openvid-1m}, which provides detailed text captions for each video. We center-crop the input training frames to a resolution of \(512 \times 512\), with a frame length of \(16\), and randomly select an anchor video as a \(16\)-frame slice from the video. We expand the input channels of the original U-Net from \(4\) to \(13\), initializing the additional convolutional layer parameters to zero. The entire U-Net was trained for 450K iterations with a learning rate of \(10^{-5}\) and a batch size of \(2\).

During inference, we process input videos exceeding \(16\) frames in length using multiple overlapping slices. To maintain temporal consistency, each subsequent slice (after the first) incorporates the last \(4\) frames from the previous slice, concatenated with the next \(12\) frames as input. This overlapping strategy ensures smooth transitions between slices and mitigates temporal discontinuities in the reconstructed video.

\subsection{Depth Scale Alignment \& View Warping Details}
\label{supp: depth_scale_alignment_and_view_warping}

\boldparagraph{Depth Scale Alignment}
To align the monocular depth predicted by the model to the metric scale, we utilize the sparse point cloud \(\{X_i\}_{i=1}^M\) from COLMAP~\cite{sfm, mvs} and apply least squares algorithm combined with the RANSAC algorithm~\cite{ransac} to achieve a more robust and accurate depth map, effectively reducing the impact of outliers and improving alignment reliability.

Specifically, given a video \(\mathcal{V} = \{I_t\}_{t=1}^T\) with \(T\) frames and its camera intrinsics \(\mathbf{K}\), camera poses \(\{\mathbf{T}_t\}_{t=1}^T\), we first project the sparse 3D points onto the image plane for each frame \(t\):
\[
\mathbf{P}_t^i = \mathbf{K} [\mathbf{T}_t \tilde{\mathbf{X}}_i]_{1:3}, \quad 
    d_t^i = [\mathbf{P}_t^i]_3, \quad 
    p_t^i = [\mathbf{P}_t^i/d_t^i]_{1:2}
\]
where \(\tilde{X}_i\) is the homogeneous coordinate of each point in world frame, \([\cdot]\) extracts the corresponding component of the coordinate. We determine valid points located in the image plane based on pixel coordinates, and then compute an optimal scale factor \(\alpha_t\) and a shift factor \(\beta_t\) using RANSAC combined with least squares optimization:

For each frame \(t\), we perform RANSAC for \(K = 100\) iterations. In each iteration:
\begin{enumerate}
    \item Randomly sample 10 valid projected points \(\{p_t^j\}_{j=1}^{10}\) with their corresponding metric depths \(\{d_t^j\}_{j=1}^{10}\).
    \item Compute candidate scale \(\alpha_t^{(k)}\) and shift \(\beta_t^{(k)}\) by minimizing the least square error:
    \[
    \alpha_t^{(k)}, \beta_t^{(k)} = \argmin_{\alpha, \beta} \sum_{j=1}^{10} \left(\alpha \hat{D}_t(p_t^j) + \beta - d_t^j\right)^2
    \]
    \item Apply the computed scale and shift to align the predicted depth map: \(D_t^{(k)} = \alpha_t^{(k)} \hat{D}_t + \beta_t^{(k)}\).
    \item Back-project all valid pixels using the aligned depth to obtain a point cloud \(\mathcal{P}_t^{(k)}\).
    \item Compute the Chamfer distance between this point cloud and the corresponding COLMAP sparse points:
    \begin{align*}
        \text{CD}_t^{(k)} = &\frac{1}{|\mathcal{P}_t^{(k)}|} \sum_{p \in \mathcal{P}_t^{(k)}} \min_{q \in \mathcal{X}_t} \|p - q\|_2^2 + \\
        &\frac{1}{|\mathcal{X}_t|} \sum_{q \in \mathcal{X}_t} \min_{p \in \mathcal{P}_t^{(k)}} \|q - p\|_2^2
    \end{align*}
    where \(\mathcal{X}_t\) represents the COLMAP sparse points visible in frame \(t\).
\end{enumerate}
Finally, we select the optimal scale and shift parameters that yield the minimum Chamfer distance:
\[
\alpha_t, \beta_t = \alpha_t^{(k^*)}, \beta_t^{(k^*)} \quad \text{where} \quad k^* = \argmin_k \text{CD}_t^{(k)}
\]
The final metric-aligned depth maps are obtained as \(D_t = \alpha_t \hat{D}_t + \beta_t\).

\boldparagraph{View Warping}
The view warping stage transforms each frame to a desired novel viewpoint using the aligned depth maps and camera poses independently. For a target camera pose \(\mathbf{T}_t'\), we perform a two-step projection process. First, we back-project each pixel \(p_t^m\) from frame \(I_t\) into 3D space using its depth value:
\[
\tilde{X}(p_t^m) = \mathbf{T}_t^{-1} \tilde{p}_{cam, t}^m, \quad 
p_{cam, t}^m = D_t(p_t^m) \mathbf{K}^{-1} \tilde{p}_t^m
\]
where \(\tilde{\cdot}\) represents the homogeneous coordinate. These 3D points are then projected onto the target view by perspective projection \(\pi(\cdot)\):
\[
{p_t^{m}}' = \pi(\tilde{X}(p_t^m), \mathbf{T}_t', \mathbf{K})
\]
The warping process inevitably creates regions without valid projections due to occlusions. This results in a masked video sequence \(\mathcal{V}' = \{I_t'\}_{t=1}^T\), where each warped frame \(I_t' \in \mathbb{R}^{H\times W\times 3}\) is accompanied by a binary mask \(\mathcal{M}_t' \in \{0,1\}^{H\times W}\) indicating valid pixels (\(1\)) and masked regions (\(0\)).

\begin{table}[t]
    \centering
    \small
    \begin{tabular}{lccc}
    \toprule
    \multicolumn{1}{l}{\textbf{Dataset}} & \multicolumn{1}{c}{\multirow{2}{*}{Start Frame}} & \multicolumn{1}{c}{\multirow{2}{*}{End Frame}} & \multicolumn{1}{c}{\multirow{2}{*}{Num Frames}}\\
    \cmidrule(lr){1-1} 
    Scene \\
    \midrule
    \multicolumn{3}{l}{\textbf{iPhone Dataset}} \\
    \midrule
    apple & 290 & 321 & 32 \\
    block & 240 & 303 & 64 \\
    paper windmill & 56 & 136 & 81 \\
    spin & 10 & 90 & 81 \\
    teddy & 185 & 265 & 81 \\
    \midrule
    \multicolumn{3}{l}{\textbf{HyperNeRF Dataset}} \\
    \midrule
    3dprinter & 40 & 103 & 64 \\
    broom & 106 & 169 & 64 \\
    chicken & 40 & 103 & 64 \\
    \bottomrule
    \end{tabular}
    \vspace{-0.25cm}
    \caption{\textbf{Details of the datasets we test for 4D Reconstrcution.}}
    \label{tab: supp_datasets}
\end{table}

\begin{table}[t]
    \centering
    \begin{tabular}{l@{\hspace{12pt}}c@{\hspace{7pt}}c@{\hspace{7pt}}c}
        \toprule
        Backbone & mPSNR~$\uparrow$ & mSSIM~$\uparrow$ & mLPIPS~$\downarrow$ \\
        \midrule
        4D GS & 16.16 & 0.4676 & 0.5164 \\
        Motion Field~(ours) & \bf\best{16.80} & \bf\best{0.5170} & \bf\best{0.4750} \\
        \bottomrule
    \end{tabular}
    \vspace{-0.25cm}
    \caption{\textbf{Quantitative ablation study on reconstruction backbones on iPhone dataset and HyperNeRF dataset.}}
    \label{tab: ablation_backbone}
\end{table}

\begin{table}[t]
    \centering
    \begin{adjustbox}{max width=\linewidth}
    \begin{tabular}{l@{\hspace{14pt}}c@{\hspace{10pt}}c@{\hspace{10pt}}c@{\hspace{12pt}}c}
    \toprule
    Method & PSNR~$\uparrow$ & SSIM~$\uparrow$ & LPIPS~$\downarrow$ & FVD~$\downarrow$ \\
    \midrule
    CoCoCo & \sbest{24.98} & \sbest{0.8213} & \sbest{0.0987} & \bf\best{12.58} \\
    StereoCrafter & \tbest{23.44} & \tbest{0.8006} & \tbest{0.1498} & 52.63 \\
    ViewCrafter & 16.95 & 0.5822 & 0.2560 & \tbest{52.07} \\
    Ours & \bf\best{27.22} & \bf\best{0.8223} & \bf\best{0.0801} & \sbest{14.30} \\
    \bottomrule
    \end{tabular}
    \end{adjustbox}
    \vspace{-0.25cm}
    \caption{\textbf{Quantitative comparison of novel view-aware video in-painting on our processed \(5K\) videos from OpenVid-1M.}}
    \label{tab: video_inpainting_comparison}
\end{table}

\begin{figure*}[t]
    \centering
    \begin{tabular}{{@{}c@{\hspace{2pt}}c@{\hspace{2pt}}c@{\hspace{2pt}}c@{\hspace{2pt}}c@{\hspace{2pt}}c@{}}}
        Shape of Motion & CoCoCo & StereoCrafter & ViewCrafter & Ours & Ground Truth \\
        \includegraphics[width=0.15\textwidth]{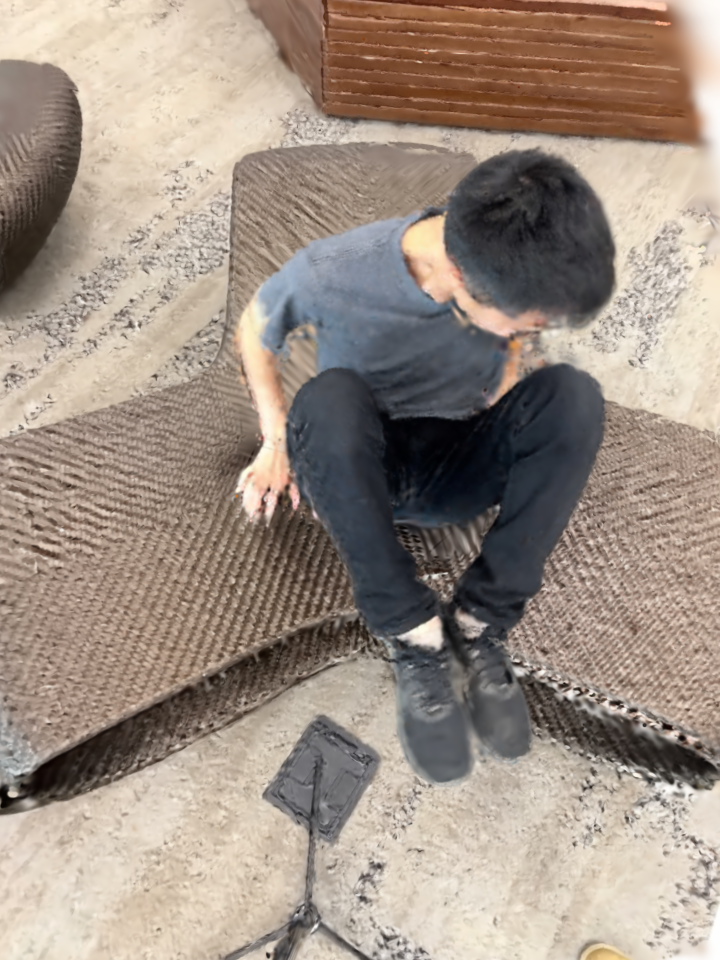} &
        \includegraphics[width=0.15\textwidth]{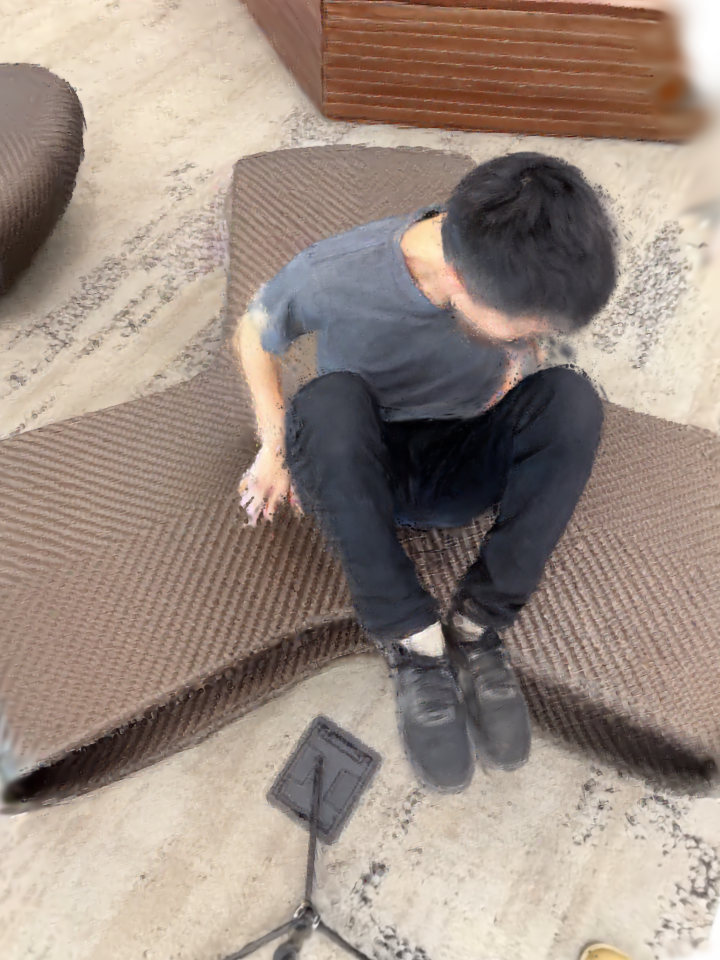} &
        \includegraphics[width=0.15\textwidth]{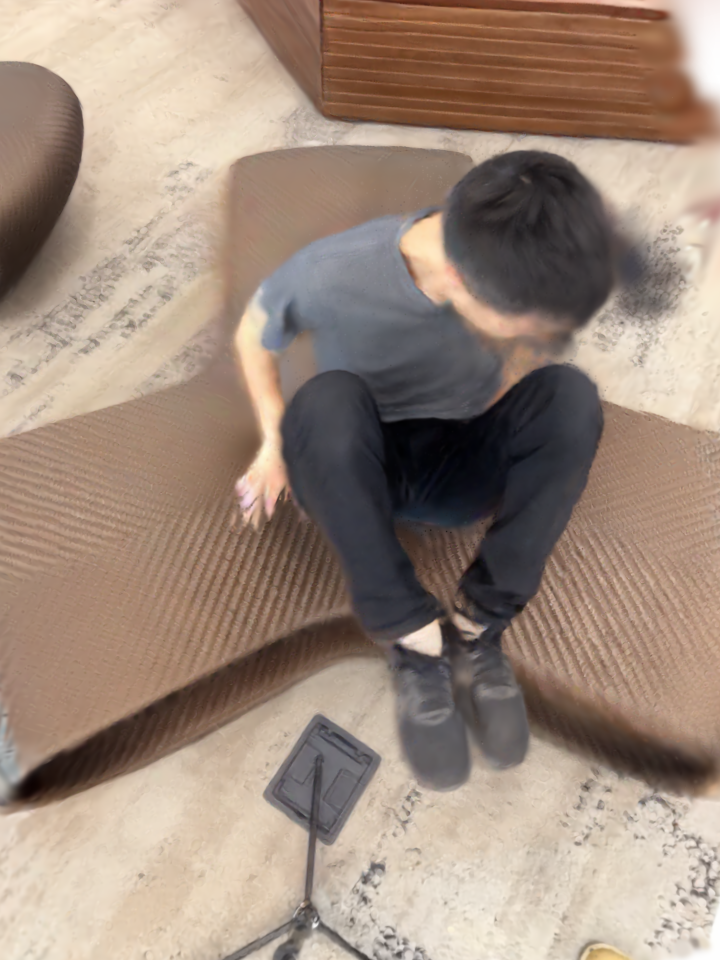} &
        \includegraphics[width=0.15\textwidth]{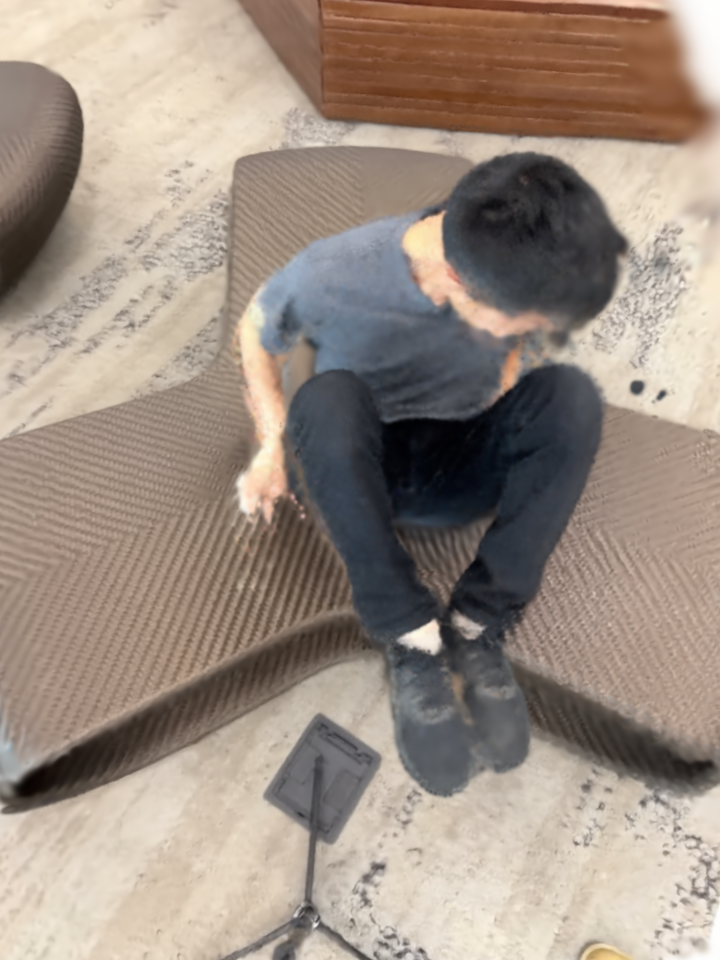} &
        \includegraphics[width=0.15\textwidth]{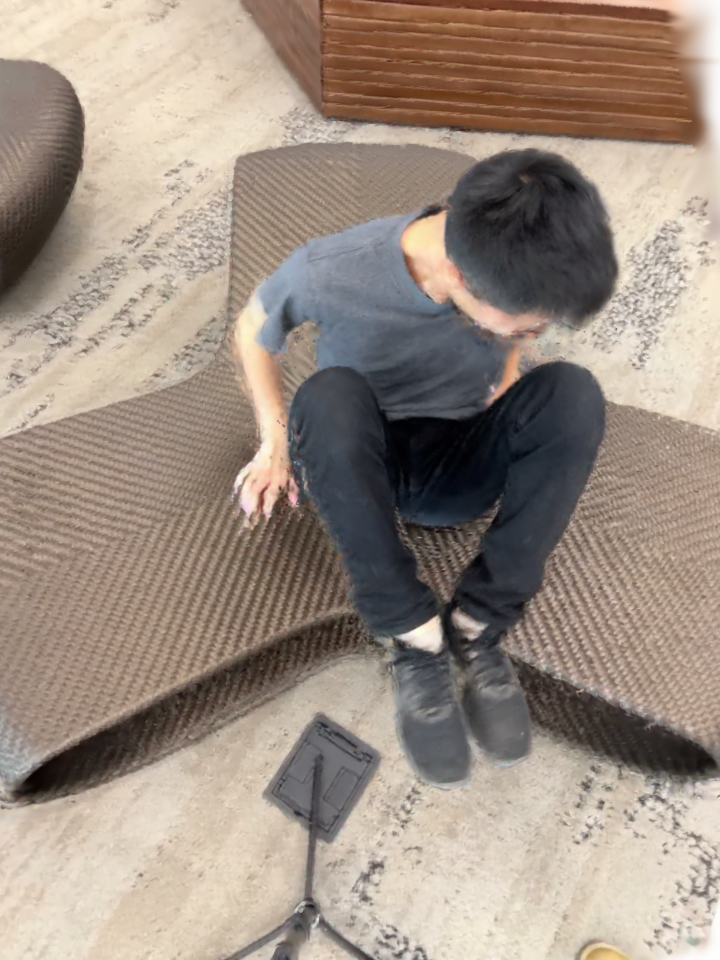} &
        \includegraphics[width=0.15\textwidth]{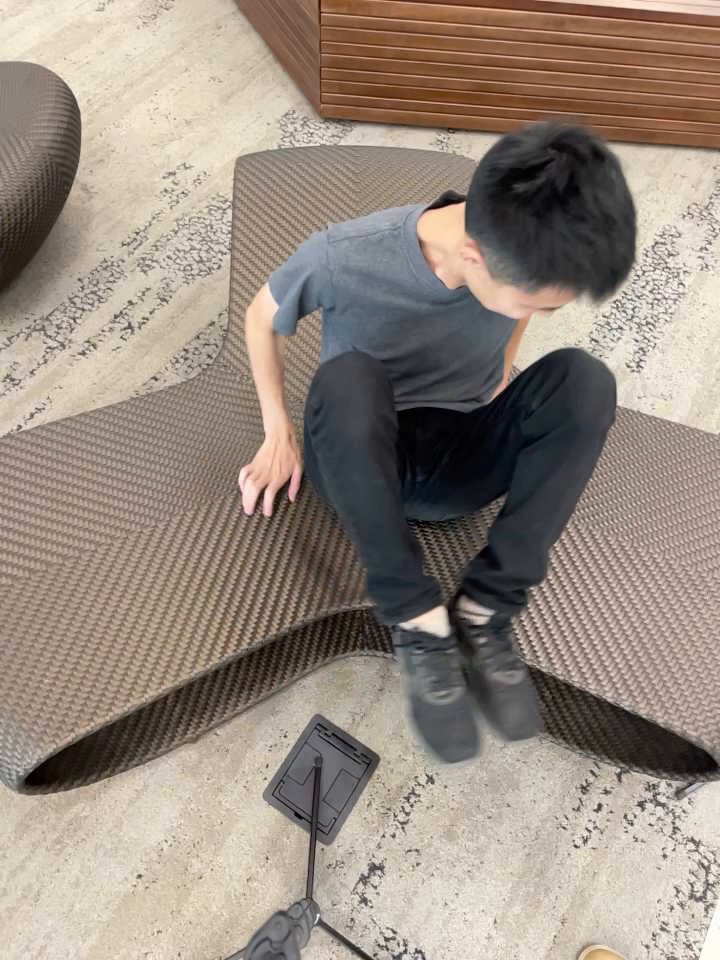} \\[-2.5pt]
        \includegraphics[width=0.15\textwidth]{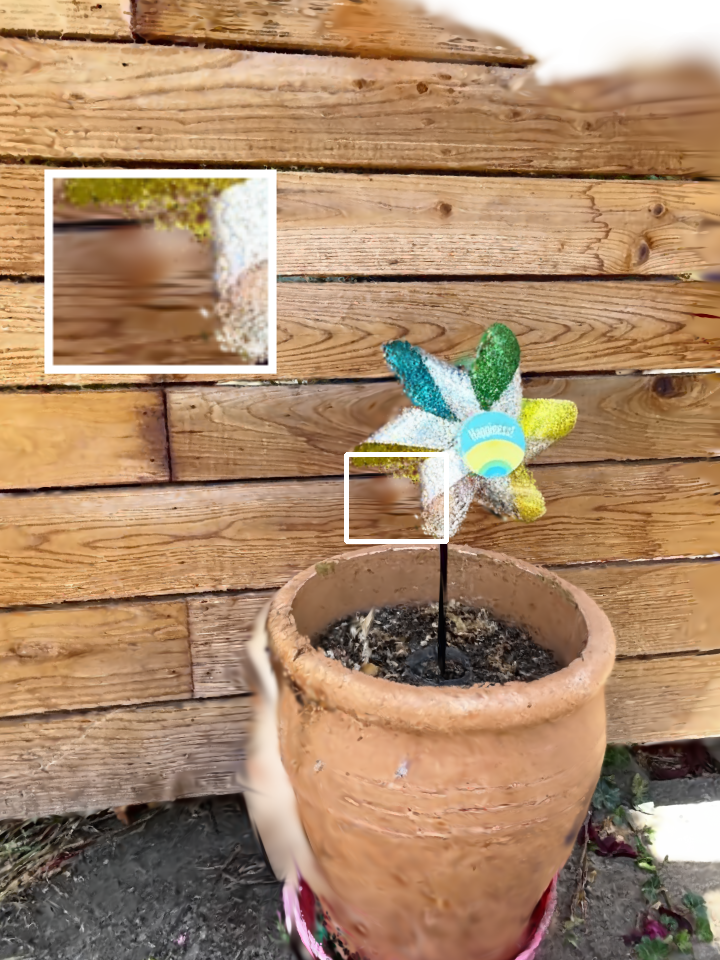} &
        \includegraphics[width=0.15\textwidth]{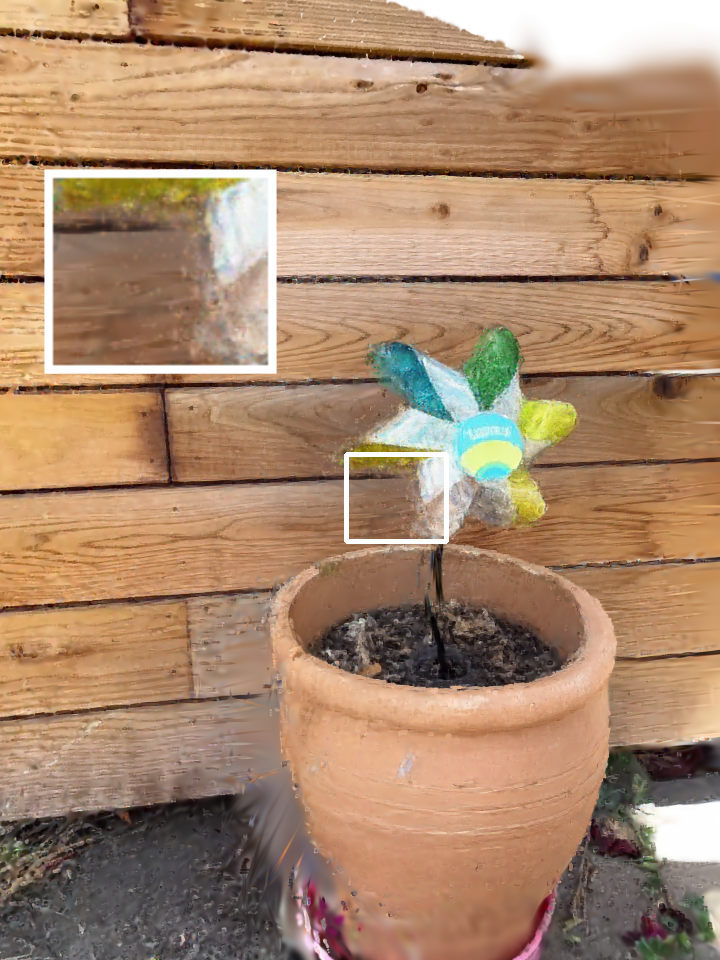} &
        \includegraphics[width=0.15\textwidth]{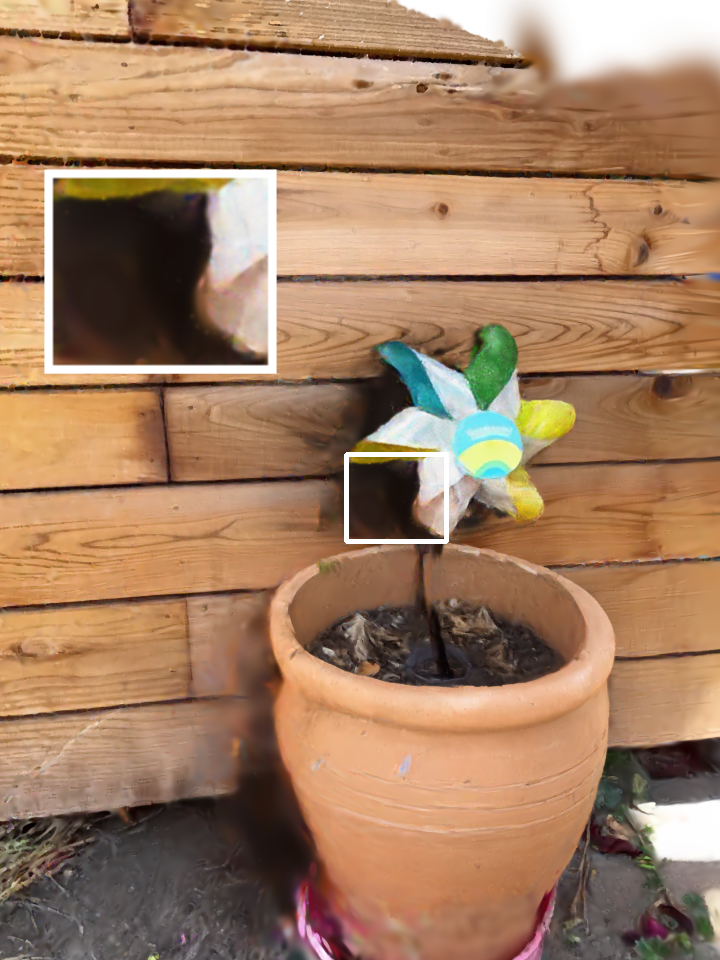} &
        \includegraphics[width=0.15\textwidth]{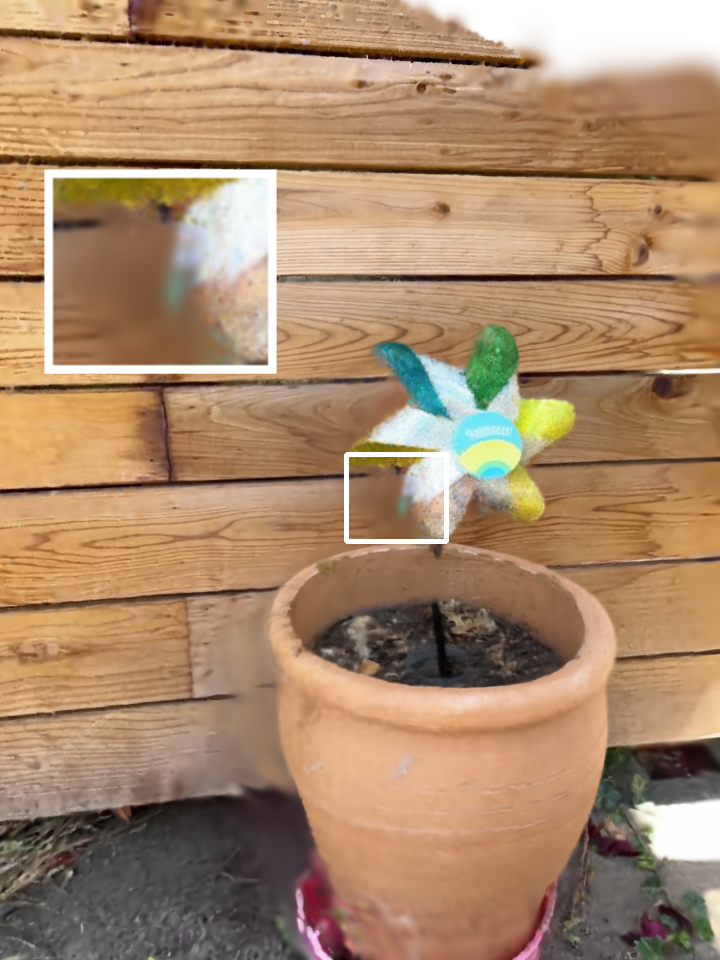} &
        \includegraphics[width=0.15\textwidth]{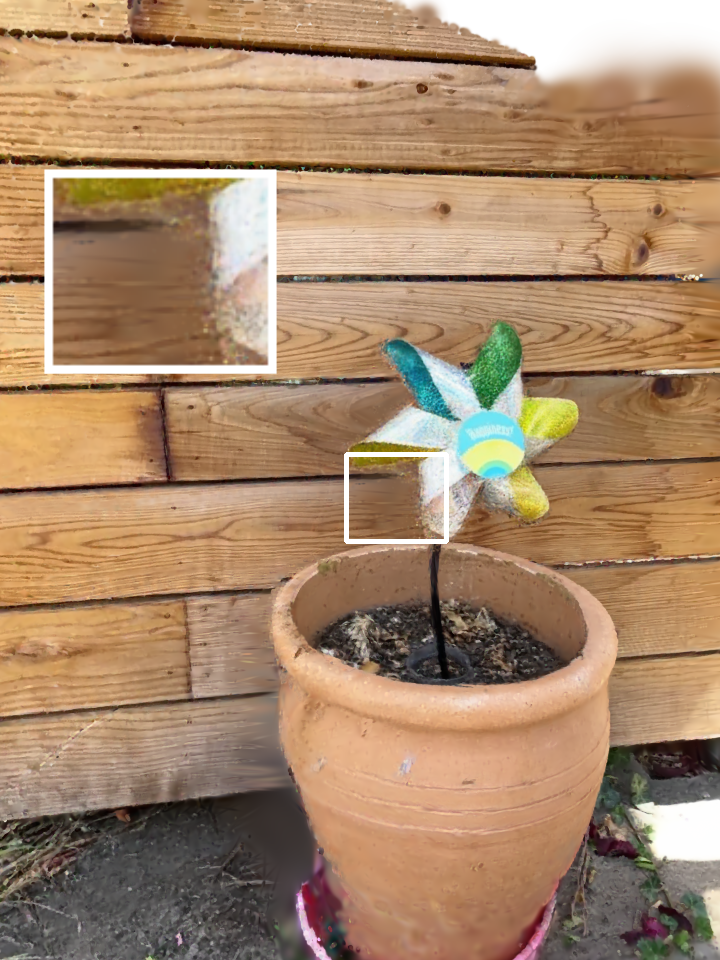} &
        \includegraphics[width=0.15\textwidth]{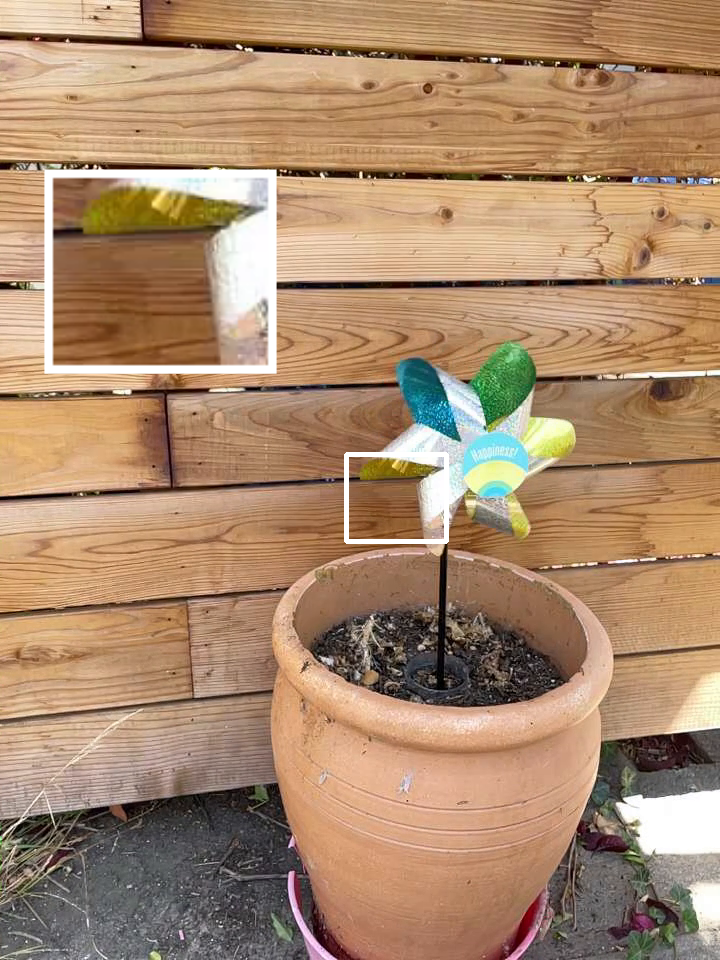} \\[-2.5pt]
        \includegraphics[width=0.15\textwidth]{figure/reconstruction_comparison/som/teddy/teddy_cropped.png} &
        \includegraphics[width=0.15\textwidth]{figure/reconstruction_comparison/cococo/teddy/teddy_cropped.png} &
        \includegraphics[width=0.15\textwidth]{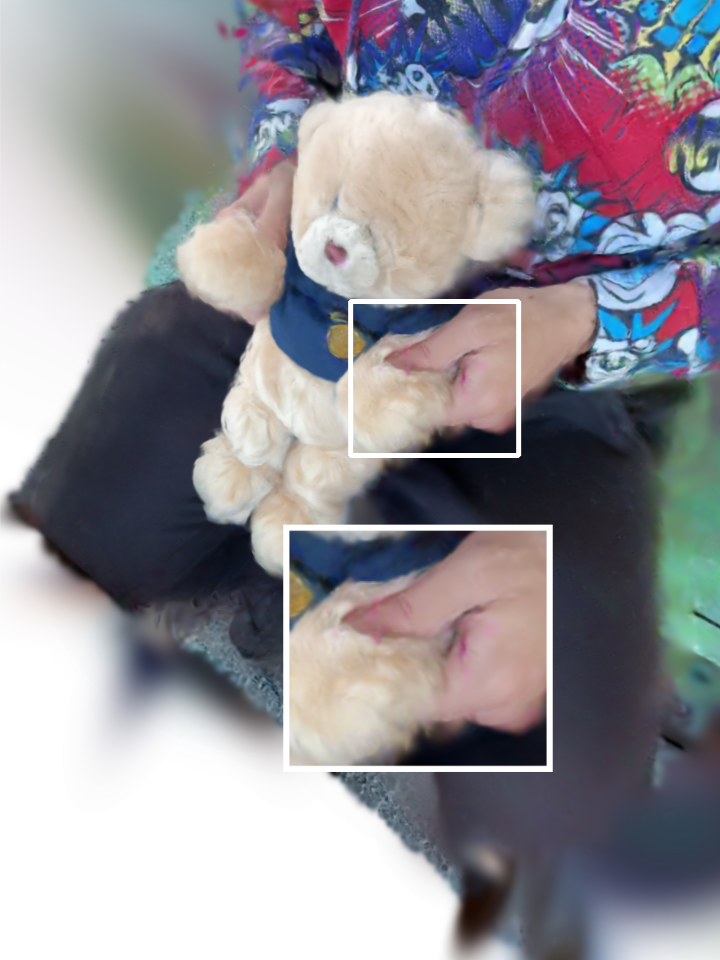} &
        \includegraphics[width=0.15\textwidth]{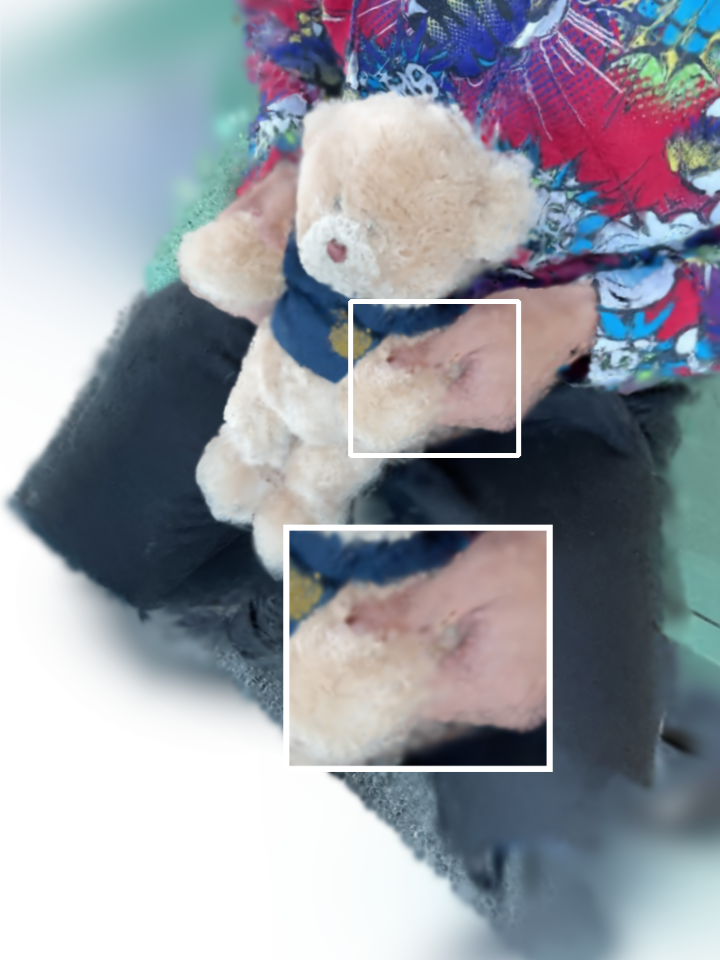} &
        \includegraphics[width=0.15\textwidth]{figure/reconstruction_comparison/ours/teddy/teddy_cropped.png} &
        \includegraphics[width=0.15\textwidth]{figure/reconstruction_comparison/gt/teddy/teddy_cropped.png} \\[-2.5pt]
        \includegraphics[width=0.15\textwidth]{figure/reconstruction_comparison/som/broom/broom_cropped.png} &
        \includegraphics[width=0.15\textwidth]{figure/reconstruction_comparison/cococo/broom/broom_cropped.png} &
        \includegraphics[width=0.15\textwidth]{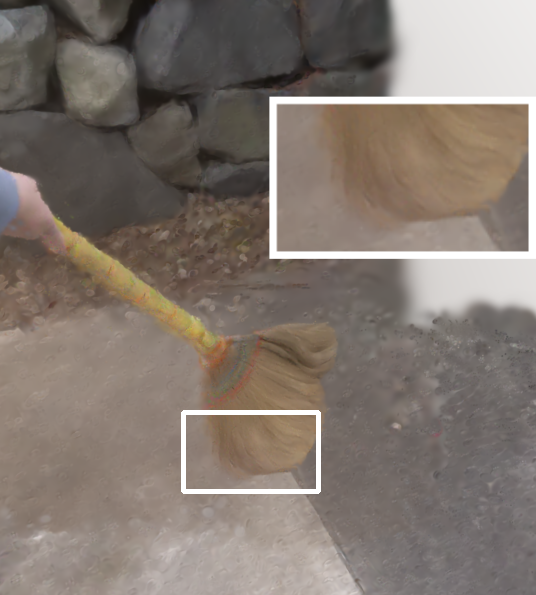} &
        \includegraphics[width=0.15\textwidth]{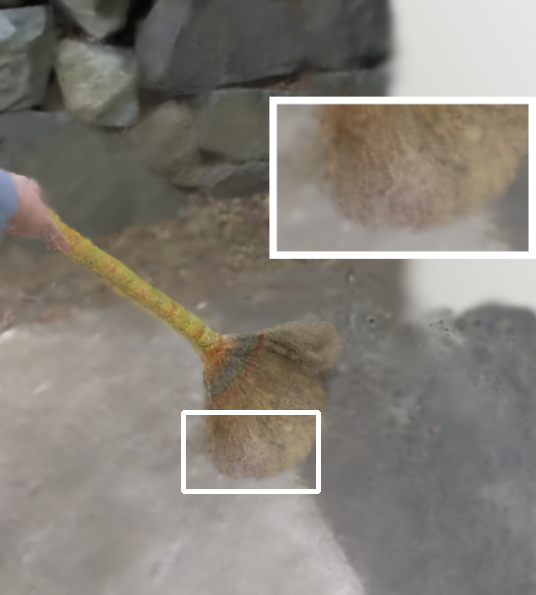} &
        \includegraphics[width=0.15\textwidth]{figure/reconstruction_comparison/ours/broom/broom_cropped.png} &
        \includegraphics[width=0.15\textwidth]{figure/reconstruction_comparison/gt/broom/broom_cropped.png}
    \end{tabular}
    \vspace{-0.25cm}
    \caption{\textbf{Qualitative comparison of dynamic scene reconstruction on iPhone dataset and HyperNeRF dataset.}}
    \label{fig: supp_reconstruction}
\end{figure*}

\subsection{Iterative View Augmentation}
\label{supp: iterave_view_augmentation}

Following recent advances in 3D reconstruction~\cite{luciddreamer, cat3d}, we pre-compute a set of target camera viewpoints in 4D space. Specifically, we start with a data buffer \(\mathcal{D}_0\) with input monocular video \(\mathcal{V}^0\) along with its aligned depths \(D^0\) and poses \(\mathbf{T}^0\), namely \(\mathcal{D}_0 = (\mathcal{V}^0, D^0, \mathbf{T}^0)\). We define \(H\) novel camera poses \(\{\mathbf{T}^i\}_{i=1}^H\) distributed around the scene and set the total number of iteration to \(N\), resulting in the synthesis of \(h=H/N\) novel videos in each iteration.

During iteration \(j (1\leq j \leq N)\), we select frames with minimal warping angles from data buffer \(\mathcal{D}_{j-1}\) at each timestamp \(t\) and warp them to target poses \(\mathbf{T}_t^{j,l}\) by our depth-guided warping pipeline, where \(l \in \{1,2,...,h\}\). Specifically, we first identify the two most peripheral frames from the data buffer \(\mathcal{D}_{j-1}\) at each timestamp \(t\), denoted as \(I_{t,1}\) and \(I_{t,2}\) with poses \(\mathbf{T}_{t,1}\) and \(\mathbf{T}_{t,2}\). We then select a total of \(h\) closest unvisited target poses from the set of remaining target poses, in which each target pose is closest to either of the peripheral frames:
\[
\{\mathbf{T}_t^{j,1}, ..., \mathbf{T}_t^{j,h}\} = \argmin_{\mathbf{T}^i \in \mathcal{U}_t} \min(d(\mathbf{T}^i, \mathbf{T}_{t,1}), d(\mathbf{T}^i, \mathbf{T}_{t,2}))
\]
where \(\mathcal{U}_t\) is the set of unvisited poses for frame \(t\), and \(d(\cdot,\cdot)\) measures the pose distance. For each selected target pose \(\mathbf{T}_t^{j,l}\) where \(l \in \{1,2,...,h\}\), we find the closer peripheral frame and warp it to the target pose using our depth-guided warping pipeline. 
Next, we organize the masked frames temporally based on warp distance rank, meaning that the \(l\)-th least distant masked frames from each timestamp are grouped together into a video. This results in \(h\) novel video sequences \(\mathcal{V}^j = \{\mathcal{V}^{j,1}, ..., \mathcal{V}^{j,h}\}\) where \(\mathcal{V}^{j,l} = \{I_t^{j,l}\}_{t=1}^T\), along with corresponding masks \(\mathcal{M}^j = \{\mathcal{M}^{j,1}, ..., \mathcal{M}^{j,h}\}\) with consistent occluded regions.

To complete the novel view synthesis for each of the \(h\) videos, we employ our trained video inpainting diffusion model \(\mathcal{F}\), which utilizes the input monocular video \(\mathcal{V}^0\) as an anchor along with a text description of the scene:
\[
\hat{\mathcal{V}}^{j,l} = \mathcal{F}(\mathcal{V}^{j,l}, \mathcal{M}^{j,l}, \mathcal{V}^0)
\]
where \(\hat{\mathcal{V}}^{j,l} = \{\hat{I}_t^{j,l}\}_{t=1}^T\) represents the final inpainted video sequence for the \(l\)-th novel video in iteration \(j\). 

After each successful inpainting iteration, we estimate the metric depth \(D^{j,l}\) for each of the \(h\) novel synthesized videos and update the data buffer \(\mathcal{D}_j = \mathcal{D}_{j-1} \cup \{(\hat{\mathcal{V}}^{j,l}, D^{j,l}, \mathbf{T}^{j,l})\}_{l=1}^h\), where \(\mathbf{T}^{j,l} = \{\mathbf{T}_t^{j,l}\}_{t=1}^T\). We repeat this process until all pre-defined poses are traversed by each frame.

Above all, we progressively build a set of multi-view observations of the scene. Since our pre-defined target poses cover the entire observation space with smooth transitions between adjacent poses, this iterative approach allows us to gradually expand viewpoint coverage, hence maintaining geometric accuracy. This strategy also helps mitigate floaters that typically arise from inaccurate depth estimates when warping at large angles. By the final iteration, we obtain data buffer \(\mathcal{D}_N\), which provides rich supervision signals for robust 4D scene reconstruction.

\begin{figure*}[t]
    \centering
    \begin{tabular}{{@{}c@{\hspace{2pt}}c@{\hspace{2pt}}c@{\hspace{2pt}}c@{\hspace{2pt}}c@{\hspace{2pt}}c@{}}}
        Masked Sequence & CoCoCo & StereoCrafter & ViewCrafter & Ours & Ground Truth \\
        \includegraphics[width=0.15\textwidth]{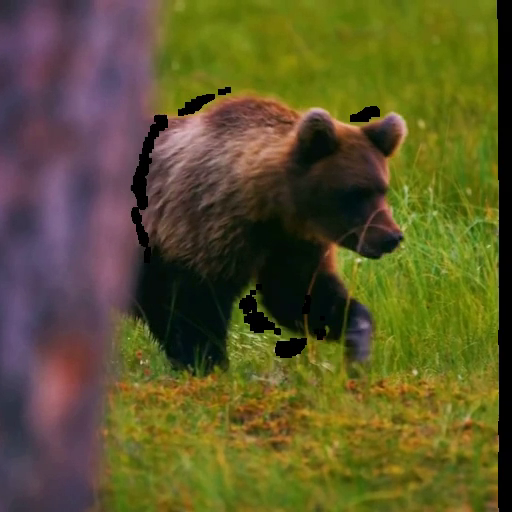} &
        \includegraphics[width=0.15\textwidth]{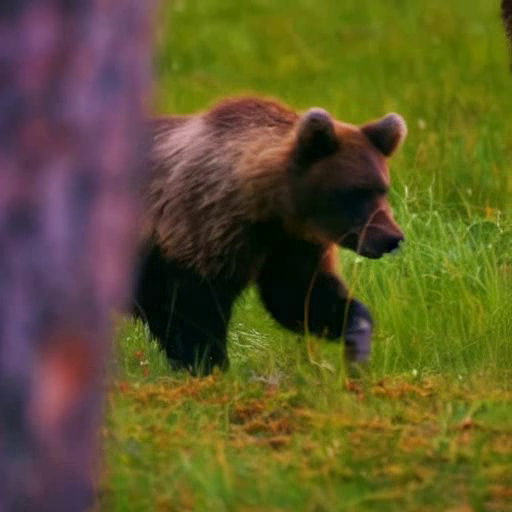} &
        \includegraphics[width=0.15\textwidth]{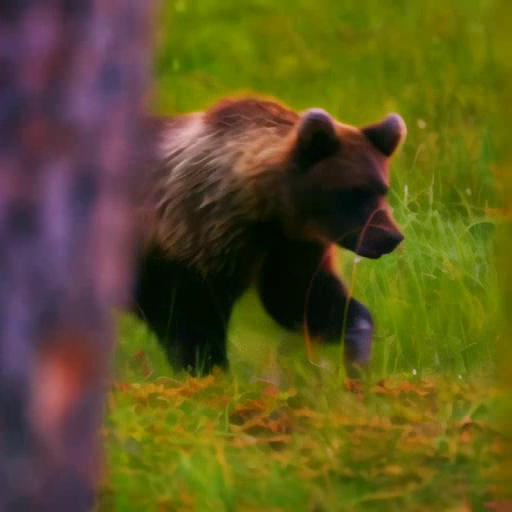} &
        \includegraphics[width=0.15\textwidth]{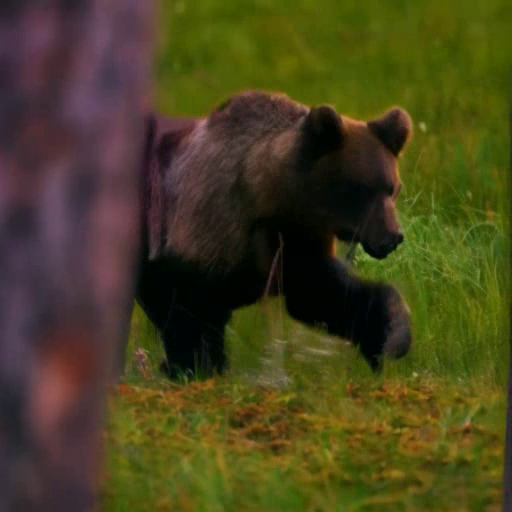} &
        \includegraphics[width=0.15\textwidth]{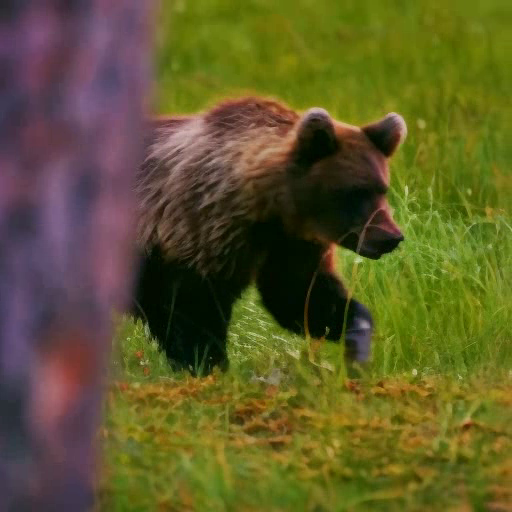} &
        \includegraphics[width=0.15\textwidth]{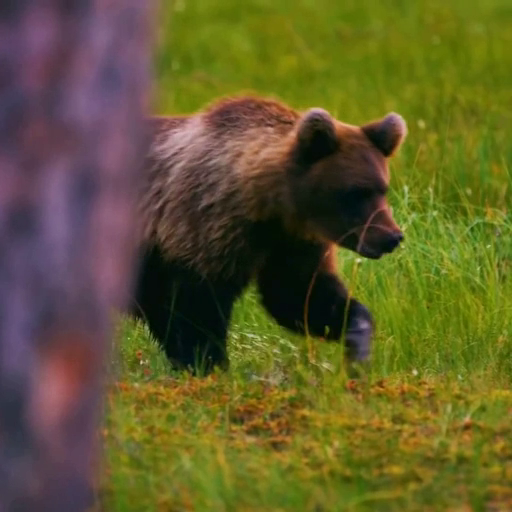} \\[-2.5pt]
        \includegraphics[width=0.15\textwidth]{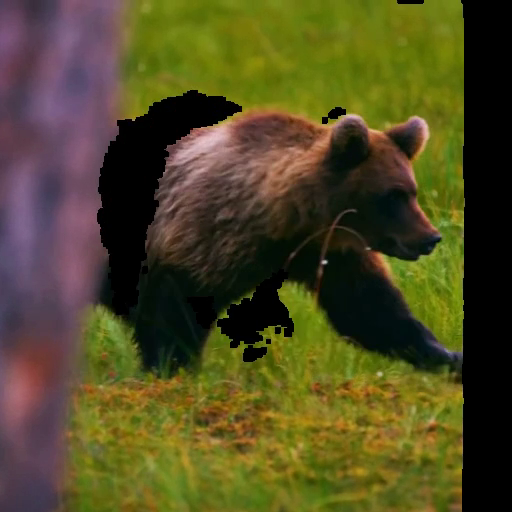} &
        \includegraphics[width=0.15\textwidth]{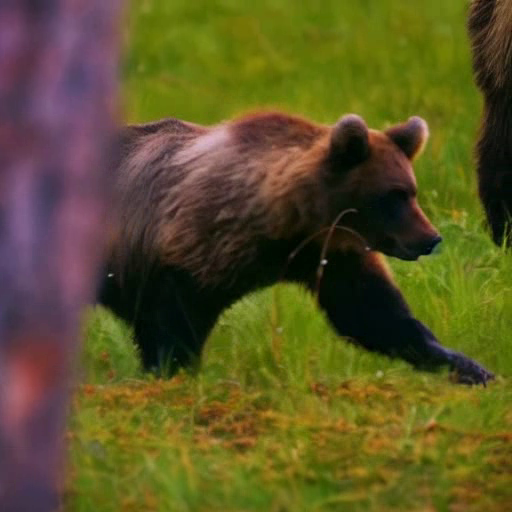} &
        \includegraphics[width=0.15\textwidth]{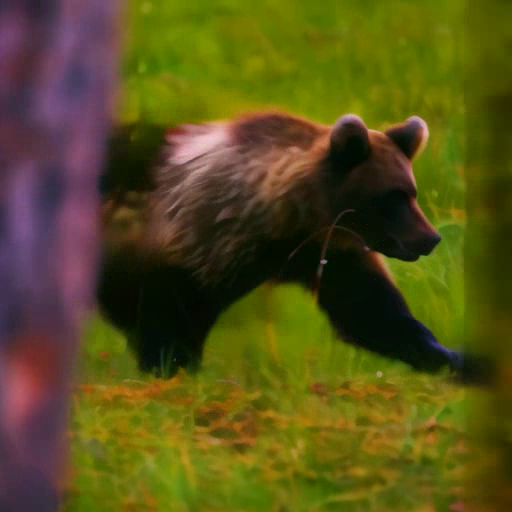} &
        \includegraphics[width=0.15\textwidth]{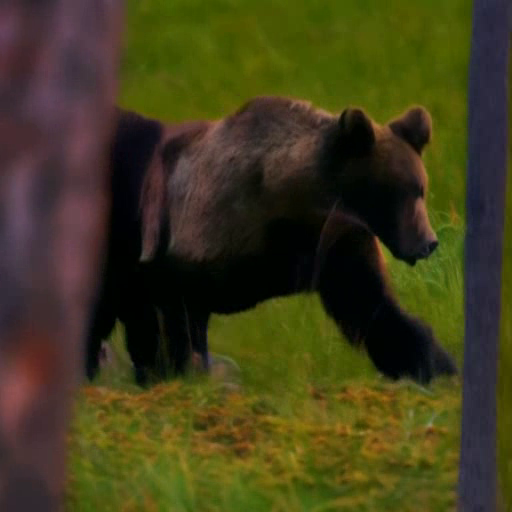} &
        \includegraphics[width=0.15\textwidth]{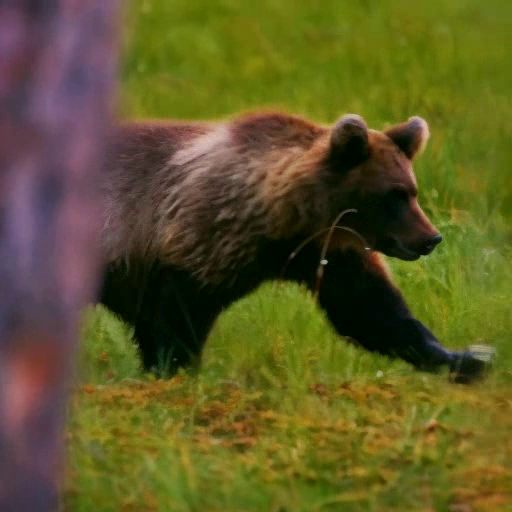} &
        \includegraphics[width=0.15\textwidth]{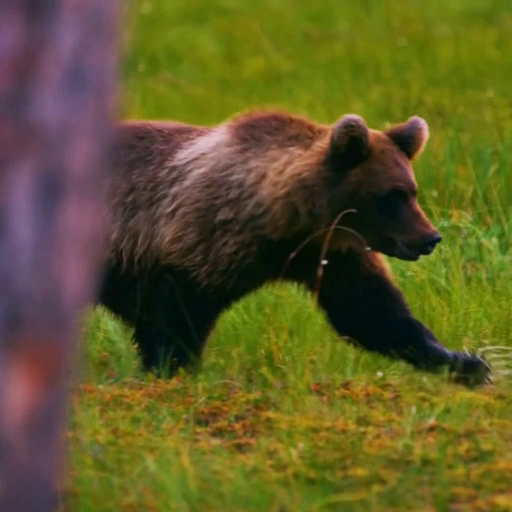} \\[-2.5pt]
        \includegraphics[width=0.15\textwidth]{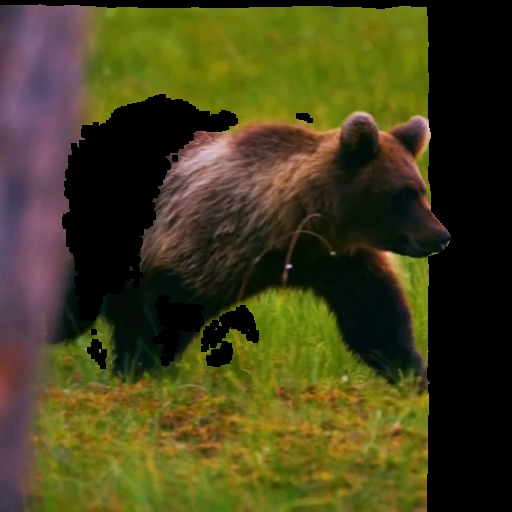} &
        \includegraphics[width=0.15\textwidth]{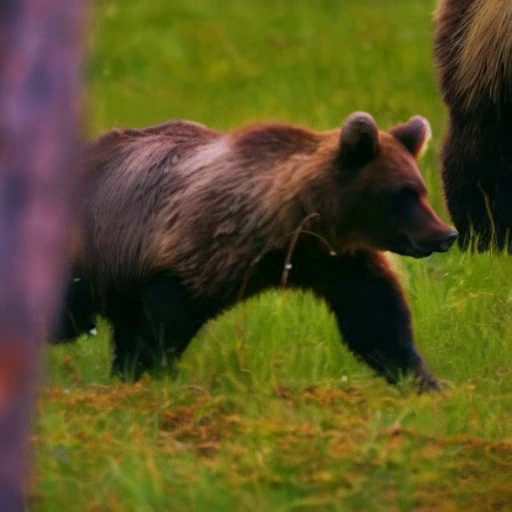} &
        \includegraphics[width=0.15\textwidth]{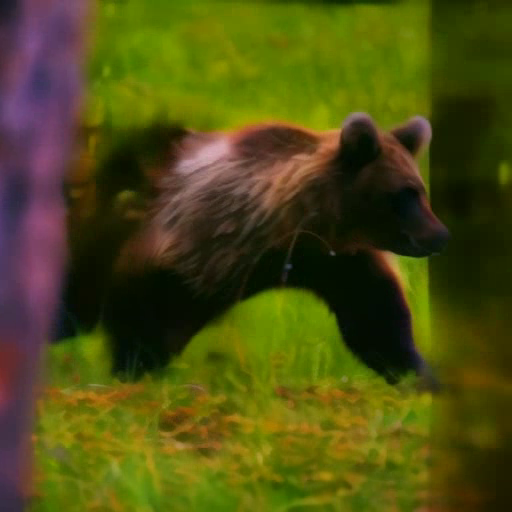} &
        \includegraphics[width=0.15\textwidth]{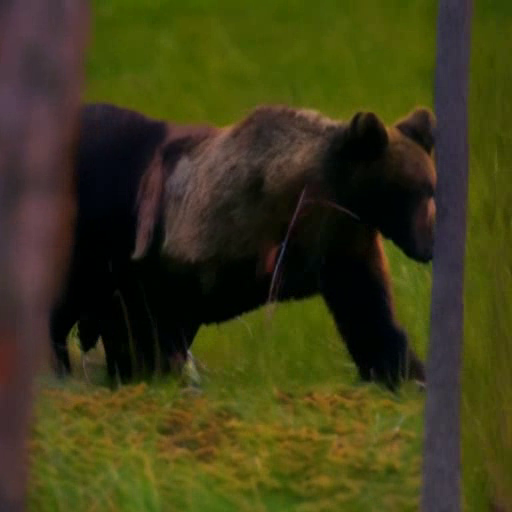} &
        \includegraphics[width=0.15\textwidth]{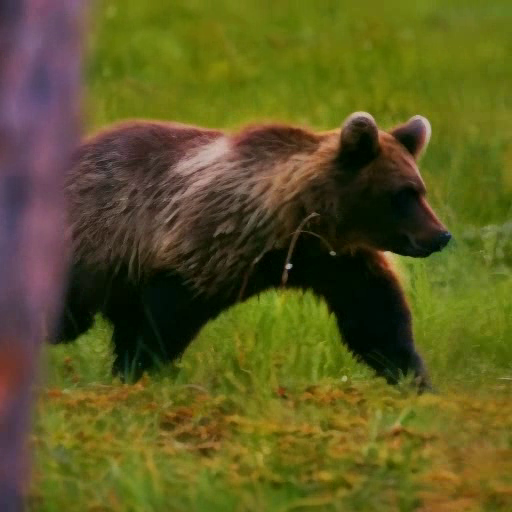} &
        \includegraphics[width=0.15\textwidth]{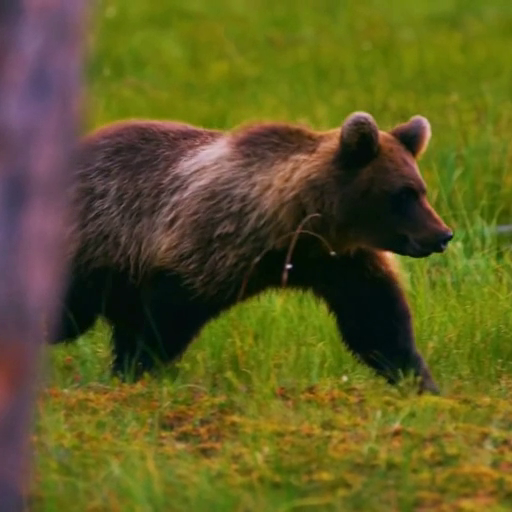}
    \end{tabular}
    \vspace{-0.25cm}
    \caption{\textbf{Qualitative comparison of novel view-aware video in-painting on our processed \(5K\) videos from OpenVid-1M.}}
    \label{fig: video_inpainting_comparison}
\end{figure*}

\begin{figure}[ht]
    \centering
    \begin{tabular}{{@{}c@{\hspace{2pt}}c@{}}}
        4D GS & Motion Field~(ours) \\
        \includegraphics[width=0.23\textwidth]{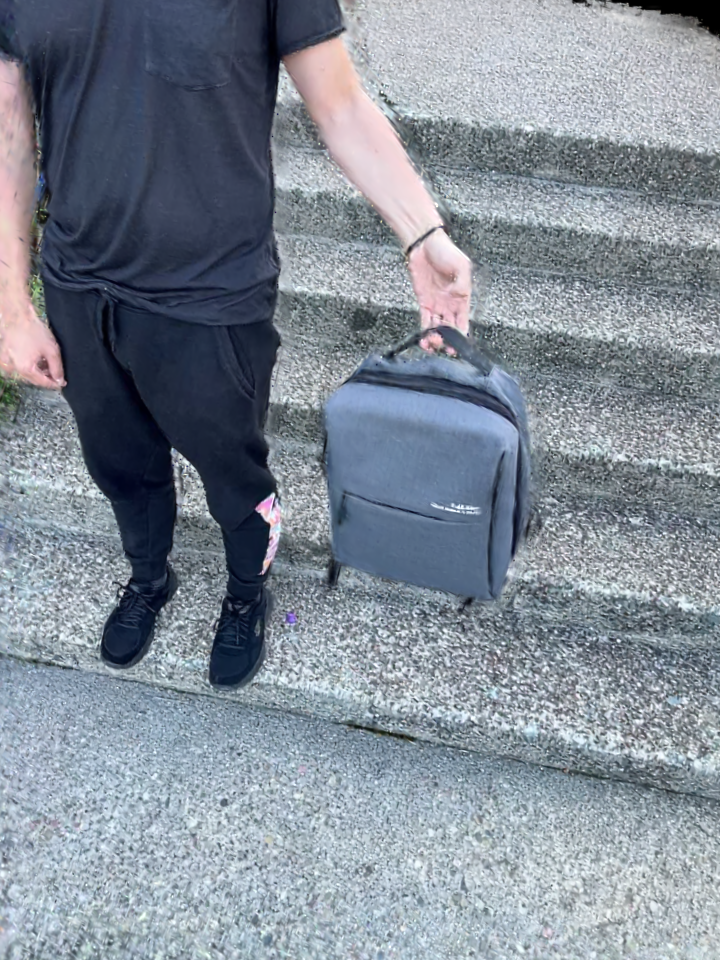} &
        \includegraphics[width=0.23\textwidth]{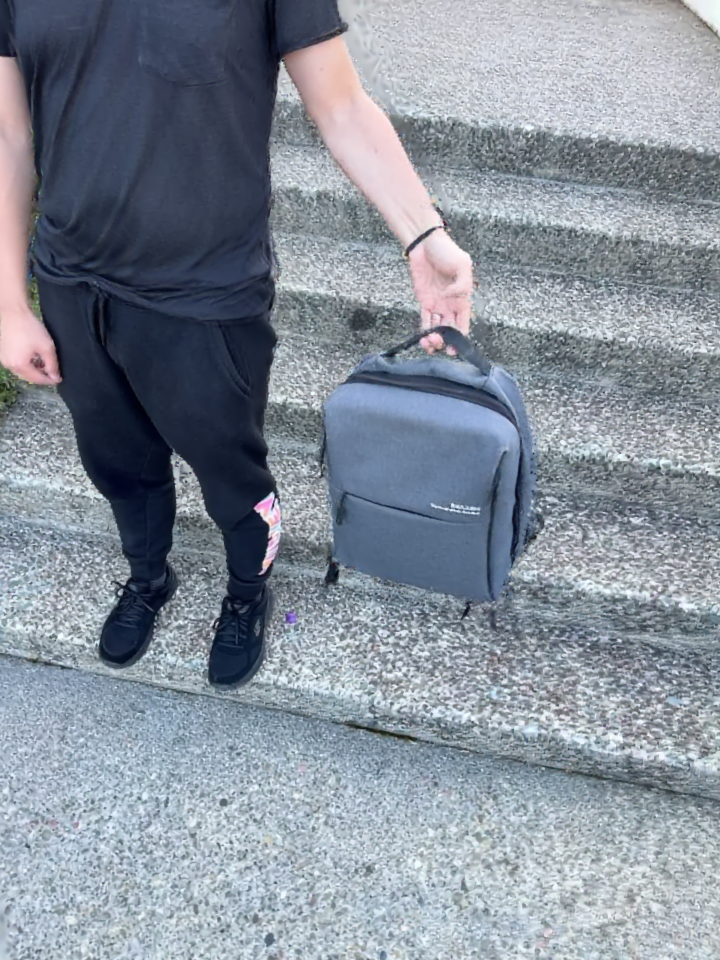} \\[-2.5pt]
        \includegraphics[width=0.23\textwidth]{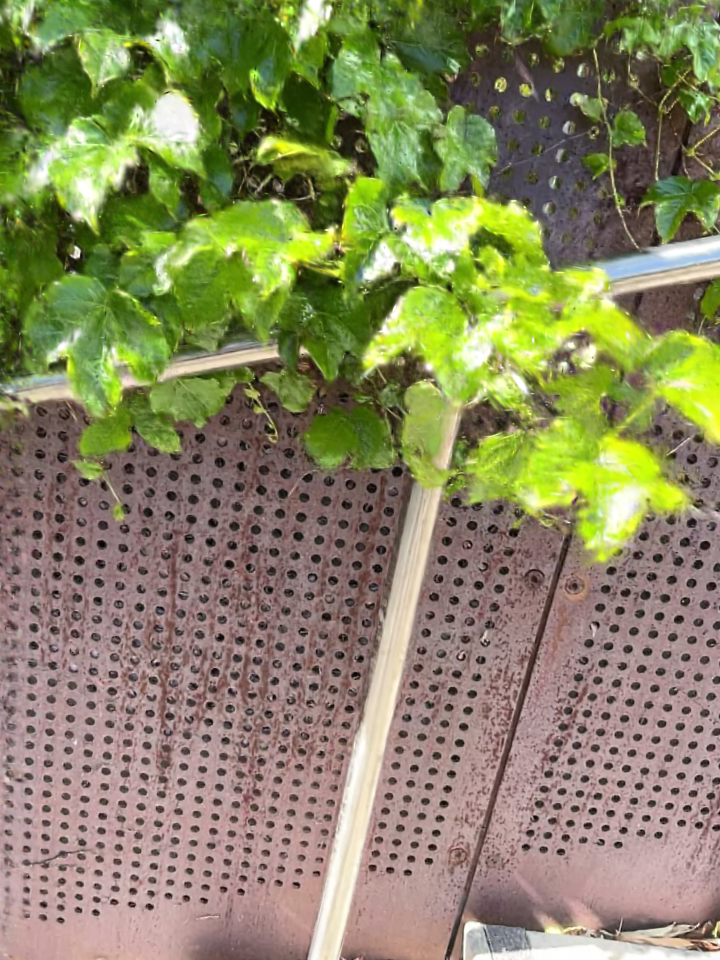} &
        \includegraphics[width=0.23\textwidth]{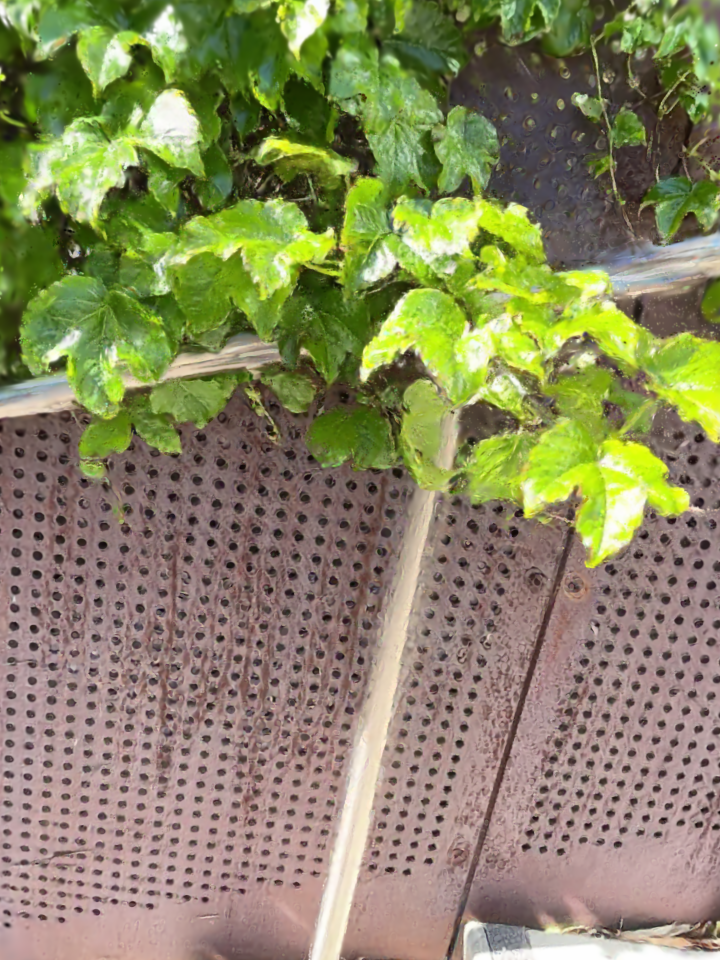}
    \end{tabular}
    \vspace{-0.25cm}
    \caption{\textbf{Qualitative ablation study on reconstruction backbones on iPhone dataset.}}
    \label{fig: ablation_backbone}
\end{figure}

\subsection{Details for Obtaining Supervision Masks}
\label{supp: supervision mask}
During iterative view augmentation, we introduce a supervision mask to avoid the inconsistency of 4D reconstruction supervision signals caused by repeated inpainting across different iterations.

To maintain this mask, we utilize a global point cloud \(\mathcal{P} = \{P_i\}_{i=1}^K\) merged from all input monocular frames to track the inpainting history. For each inpainted frame, we first render the point cloud to the corresponding camera pose, generating a visibility mask \(V_t^{j,l}\) where \(V_t^{j,l}(p) = 1\) indicates visible regions and \(V_t^{j,l}(p) = 0\) indicates invisible regions. When a masked frame is inpainted, we check the visibility of each filled pixel \(p \in \mathcal{M}^{j,l}_t\), where \(\mathcal{M}^{j,l}_t\) is the binary mask corresponding to the masked frame. If the pixel is not yet visible (\(V_t^{j,l}(p) = 0\)), we back-project it into the global point cloud as \(P_p \in \mathbb{R}^3\)~(making this region visible in future renderings) and set its supervision mask value \(S_t^{j,l}(p) = 1\). If the pixel is already visible (\(V_t^{j,l}(p) = 1\)), we set \(S_t^{j,l}(p) = 0\). Note that the supervision mask for the initially visible areas of the scene is always set to \(1\). This mechanism ensures that only the first inpainted result for each region is used as supervision signal, preventing conflicting supervision from repeated inpainting of the same region.

\section{Additional Experimental Results}
\label{supp: experiments}
\subsection{Monocular 4D Reconstruction}
\label{supp: monocular 4D Reconstruction}

\boldparagraph{Datasets}
We select a continuous clip containing a few dozen frames from each test scene in the iPhone~\cite{dycheck} and HyperNeRF~\cite{hypernerf} datasets, rather than using all available observations. The perspectives used for reconstruction exhibit significant parallax compared to the test viewpoints in these cases. The specific scenes, along with their start and end frames, are detailed in~\cref{tab: supp_datasets}.

\boldparagraph{Results}
We presented some of our 4D reconstruction results in the main paper~(~\cref{sec: scene reconstruction}). Here, we further provide a qualitative comparison with StereoCrafter~\cite{stereocrafter} and ViewCrafter~\cite{viewcrafter} in~\cref{fig: supp_reconstruction} to illustrate our superiority in reconstruction quality. It can be observed that StereoCrafter produces results with noticeable color differences, which is primarily due to limitations in its video inpainting model when handling large missing areas. ViewCrafter, while avoiding color discrepancies, also suffers from blurry outputs, as the input masked videos contain dynamic objects. In contrast, our method generates significantly clearer results with minimal deviation from the ground truth, demonstrating the superior capability of our video inpainting model in generating warped multi-view videos.

\subsection{Novel View-Aware Video inpainting}
\label{supp: video inpainting}

\boldparagraph{Dataset and Metrics} 
We compare the novel view-aware video inpainting task on our processed non-training \(5K\) masked videos from OpenVid~\cite{openvid-1m}, where an anchor video is paired with each masked video. We employ PSNR, SSIM~\cite{ssim}, LPIPS~\cite{lpips} and FVD~\cite{fvd} as the evaluation metrics for assessing inpainting quality. 

\boldparagraph{Comparison Baselines} 
We choose CoCoCo~\cite{cococo}, StereoCrafter~\cite{stereocrafter} and ViewCrafter~\cite{viewcrafter} as our baselines. CoCoCo is a text-guided video inpainting diffusion model with strong consistency and controllability. StereoCrafter trains a video diffusion model for stereo video generation. Although ViewCrafter is a 3D-aware video inpainting method, we include it as a strong baseline due to its state-of-the-art performance in geometry-aware scene completion. For StereoCrafter, we provide the masked video and corresponding mask as input. For CoCoCo, we additionally supply a text prompt to guide the inpainting process, following its original setup. For ViewCrafter, we use the first frame from our anchor video as the image condition.

\boldparagraph{Results} 
The results are presented in~\cref{tab: video_inpainting_comparison} and~\cref{fig: video_inpainting_comparison}. Among the baseline methods, CoCoCo demonstrates superior inpainting performance due to its integration with base image generation models. However, this approach tend to generates content that deviates from the scene context. In regions with weak contextual constraints, the results exhibit significant discrepancies from the ground truth, suggesting a tendency toward generation rather than faithful reconstruction. This observation is further supported by our method's higher PSNR/SSIM/LPIPS scores in quantitative evaluations. StereoCrafter exhibits limitations in handling large-scale holes due to its task orientation and training data constraints. Viewcrafter, originally designed for 3D inpainting, shows notable distortions and color inconsistencies when handling videos with dynamic objects.

In contrast, our method produces results that are most consistent with the ground truth, demonstrating that our model effectively learns the spatiotemporal priors embedded in the anchor video. Additionally, leveraging an anchor video as a constraint provides crucial spatiotemporal priors and partial observations of occluded regions~(due to dynamic foreground objects). This additional source of information enables more accurate inpainting and \textit{reconstruction}, making our approach particularly well-suited for 4D pseudo-ground-truth generation. By incorporating natural geometric and temporal constraints, our method ensures that the inpainted results closely match the original video and maintain consistency across different views. Furthermore, subsequent 4D reconstruction experiments validate the necessity of our model design.

\subsection{Ablation Study}
\label{supp: ablation_study}

\boldparagraph{Effect of 4D Reconstruction Backbones} 
We also evaluate the reconstruction backbone of our method. The results are illustrated in~\cref{tab: ablation_backbone} and~\cref{fig: ablation_backbone}. With the supervision of multiview synchronized video, the results of 4D GS~\cite{4dgs_time} have improved significantly but still do not reach the same level as motion field~\cite{som}. This discrepancy may be attributed to differences in their representations. The combination of the canonical GS with a motion field may offer greater robustness than extending 3D GS with an additional time dimension when using generated videos as supervision.

\begin{figure}[t]
    \centering
    \begin{tabular}{{@{}c@{\hspace{4pt}}c@{}}}
        \includegraphics[width=0.48\columnwidth]{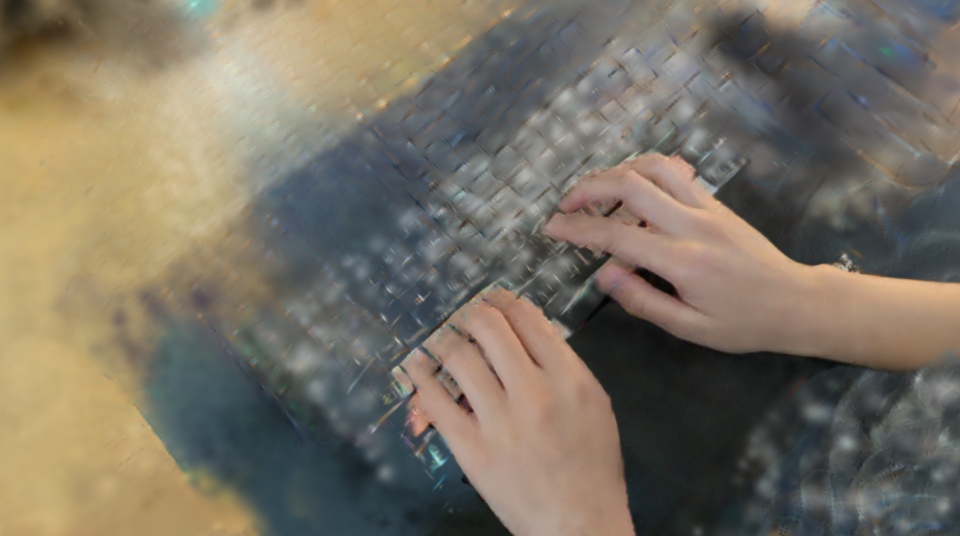} &
        \includegraphics[width=0.48\columnwidth]{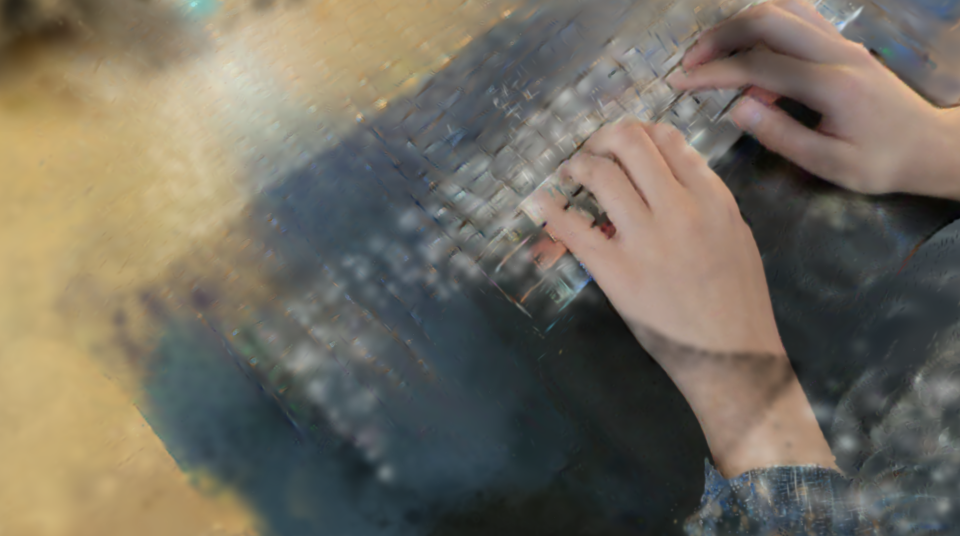}
    \end{tabular}
    \vspace{-0.25cm}
    \caption{\textbf{Failure Cases.} Our method degenerates when the depth of the input video is not correctly obtained.}
   \label{fig: failure cases}
\end{figure}

\section{Limitations}
\label{supp: limitations}

We present some failure cases in~\cref{fig: failure cases}. The main limitation of our proposed method is its dependency on the accuracy of poses and  metric depth estimation. Although we utilize robust depth scale alignment to improve depth accuracy, as well as iterative view augmentation and IV loss to mitigate the impact of depth noise on warping and reconstruction, our results may still degrade in scenes where the monocular depth prediction is highly inaccurate.

In future work, we could combine our method with some recent dense reconstruction methods~\cite{monst3r, cut3r, megasam} to obtain more accurate scene geometry, as well as improving our diffusion models by improving model architecture to better synthesize fine details during the inpainting process. Additionally, extending our approach to handle more extreme camera motions and complex scene dynamics would further enhance its applicability to a wider range of scenarios.

\end{document}